\pdfoutput=1
\documentclass[11pt]{article}

\usepackage[]{acl}
\usepackage{times}
\usepackage{latexsym}
\usepackage[T1]{fontenc}
\usepackage[utf8]{inputenc}
\usepackage{microtype}
\usepackage{inconsolata}

\usepackage[whole]{bxcjkjatype}
\usepackage{graphicx}
\usepackage{amsmath}
\usepackage{amsfonts}
\usepackage{multirow}
\usepackage{booktabs}
\usepackage{adjustbox}
\usepackage{subcaption}
\usepackage[many]{tcolorbox}
\usepackage{xcolor}
\usepackage{listings}

\newcommand{\ml}[1]{\multicolumn{1}{l}{#1}}

\newcommand{\sisj}[1]{$(S_i, S_j)$}
\newcommand{\Esisj}[1]{$\mathrm{E}(S_{i}^2 S_{j}^2)$}

\definecolor{hblue}{RGB}{30,144,255}
\definecolor{hred}{RGB}{255,0,0}

\title{Understanding Higher-Order Correlations Among\\Semantic Components in Embeddings}

\author{
Momose Oyama${}^{1,2}$\quad Hiroaki Yamagiwa${}^{1}$\quad Hidetoshi Shimodaira${}^{1,2}$\\
${}^{1}$Kyoto University\quad
${}^{2}$RIKEN\\
\texttt{oyama.momose@sys.i.kyoto-u.ac.jp},
\texttt{hiroaki.yamagiwa@sys.i.kyoto-u.ac.jp},\\
\texttt{shimo@i.kyoto-u.ac.jp}
}

\begin{document}
\maketitle
\begin{abstract}
Independent Component Analysis (ICA) offers interpretable semantic components of embeddings.
While ICA theory assumes that embeddings can be linearly decomposed into independent components, real-world data often do not satisfy this assumption. Consequently, non-independencies remain between the estimated components, which ICA cannot eliminate. We quantified these non-independencies using higher-order correlations and demonstrated that when the higher-order correlation between two components is large, it indicates a strong semantic association between them, along with many words sharing common meanings with both components. The entire structure of non-independencies was visualized using a maximum spanning tree of semantic components. These findings provide deeper insights into embeddings through ICA.
\end{abstract}

\section{Introduction}
\begin{figure}[ht]
    \centering
    \includegraphics[width=0.86\columnwidth]{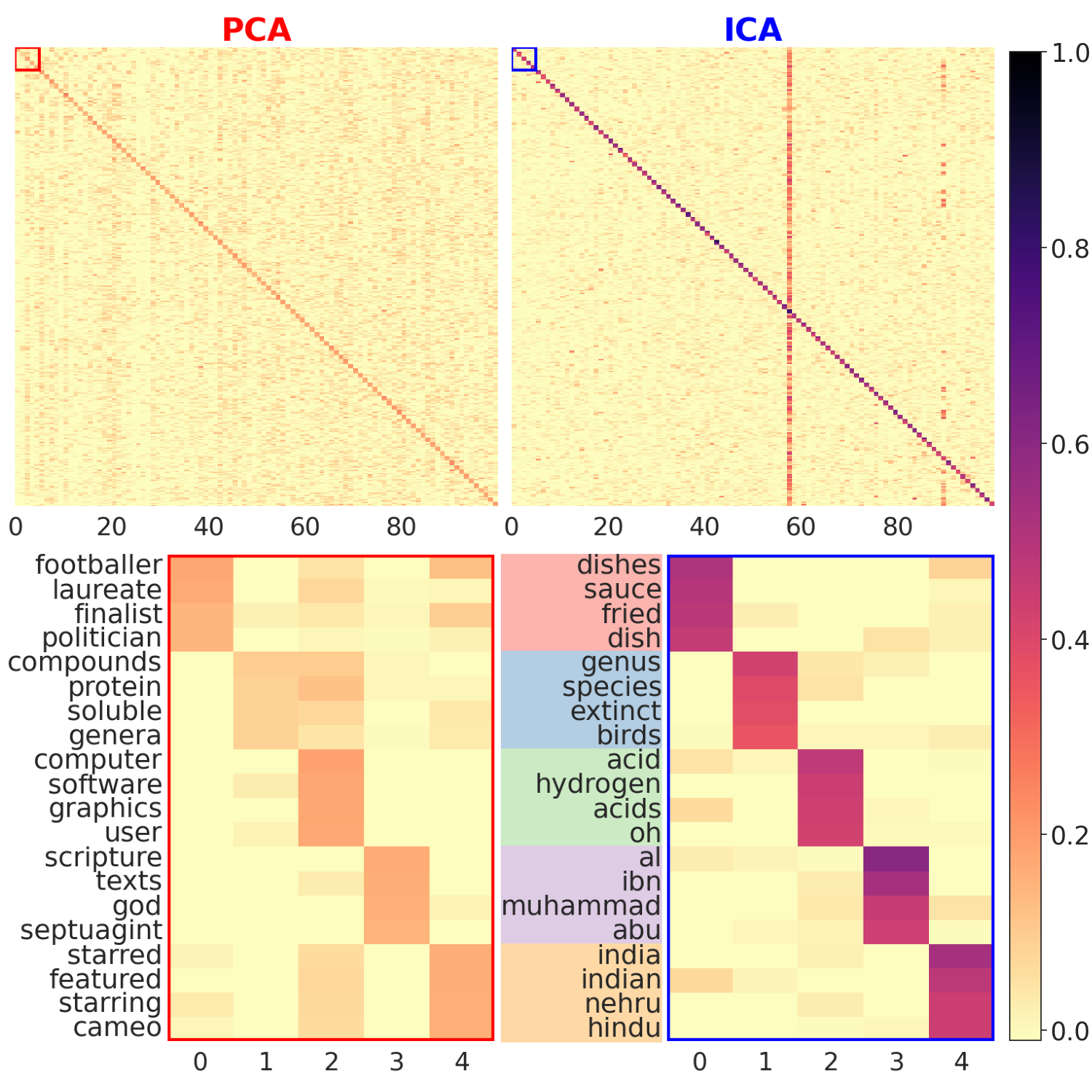}
    \caption{Heatmap visualization of 300-dimensional SGNS embeddings transformed by PCA and ICA, with
    axes sorted by variance and skewness, respectively. Each embedding has been normalized to have a norm of 1 for better visual interpretation. For each axis, the top 4 words (frequency $n_w \geq 100$ in text8) with largest component values were used.
    The first 100 axes are displayed in the top panels, and the first 5 axes with the word labels are displayed in the bottom panels. See Appendices~\ref{appendix:settings}, \ref{app:axis57} and \ref{appendix:experiment_results} for details.}
    \label{fig:heatmap_ica-vs-pca}
\end{figure}

Embeddings play an important role in natural language processing, ranging from word embeddings~\cite{sgns} to internal representations in language models~\cite{devlin-etal-2019-bert, gpt3, touvron2023llama}. 
Understanding how embeddings represent meaning is crucial for unraveling black box NLP models.

Independent Component Analysis (ICA) \cite{hyvarinen2000independent} is an effective method for visualizing and interpreting the geometric structure of embeddings~\cite{musil-marecek-2024-exploring, yamagiwa-etal-2023-discovering}.
Just as PCA aims to make coordinate axes uncorrelated, ICA seeks to transform the coordinate axes into statistically independent components.
The resulting axes from ICA tend to have sparser component values with a few larger values compared to PCA, which increases interpretability as the axes can be seen as specific semantic components (Fig.~\ref{fig:heatmap_ica-vs-pca}).

However it has been pointed out that the estimated `independent components' are only approximately independent~\cite{hyvarinen01topographic, sasaki13correlated, sasaki14estimating}.
This is because many real-world datasets cannot be accurately represented as a linear combination of independent components, contradicting the assumption of ICA theory.

In this study, we aim to further interpret the results of applying ICA to embeddings by focusing on the non-independence between `independent components'.
We measure the degree of non-independence by calculating higher-order correlations between components and find that components with large higher-order correlations can be interpreted as having strong semantic associations.
The entire structure is revealed by visualizing the maximum spanning tree of semantic components with higher-order correlations as edge weights.

\section{Review: ICA-Transformed Embeddings} \label{sec:ica}

\paragraph{Procedure of ICA.}
For a centered embedding matrix $\mathbf{X}\in \mathbb{R}^{n\times d}$ that represents the meanings of $n$ words by $d$-dimensional vectors, ICA\footnote{For the computation, \texttt{FastICA}~\cite{hyvarinen1999fast} implemented in \texttt{scikit-learn}~\cite{sklearn} is used.} seeks a transformation $\mathbf{S} = \mathbf{X}\mathbf{B}$ such that each component $S_1, \cdots, S_d$ of the transformed matrix $\mathbf{S} = [S_1, \cdots, S_d]$ is as statistically independent as possible\footnote{The $k$-th component $S_k$ is also referred to as Axis $k$.}.
The transformation $\mathbf{B}$ can be expressed as the product of the whitening transformation matrix $\mathbf{A}$ (e.g., PCA transformation) and the orthogonal transformation matrix $\mathbf{R}_{\mathrm{ica}}$, i.e., the resulting $\mathbf{S}$ is represented as
\begin{equation} \label{eq:ica_XAR}
\mathbf{S} = \mathbf{X}\mathbf{A}\mathbf{R}_{\mathrm{ica}}.
\end{equation}
Here, $\mathbf{R}_{\mathrm{ica}}$ is obtained by minimizing the mutual information\footnote{$H(X) = - \int P_{X}(x) \log P_{X}(x) dx$ is the entropy.} $I(S_1 \cdots S_d) = \sum H(S_i) - H(S_1 \cdots S_d)$, which is equivalent to maximizing the non-gaussianity\footnote{
The degree to which a probability distribution deviates from a Gaussian distribution can be measured using statistics based on higher-order moments, such as skewness (the third moment) or the negentropy of the distribution.
} of the distributions of $S_i$~\cite{hyvarinen2000independent}.
The normalized ICA-transformed embeddings, with each embedding in $\mathbf{S}$ rescaled to a norm of 1, offer high interpretability~\cite{yamagiwa-etal-2023-discovering,yamagiwa2024revisit} and are used for visualizations in this paper.

\paragraph{Comparison of PCA and ICA.}

\begin{figure}[ht]
    \centering
    \includegraphics[width=0.95\columnwidth]{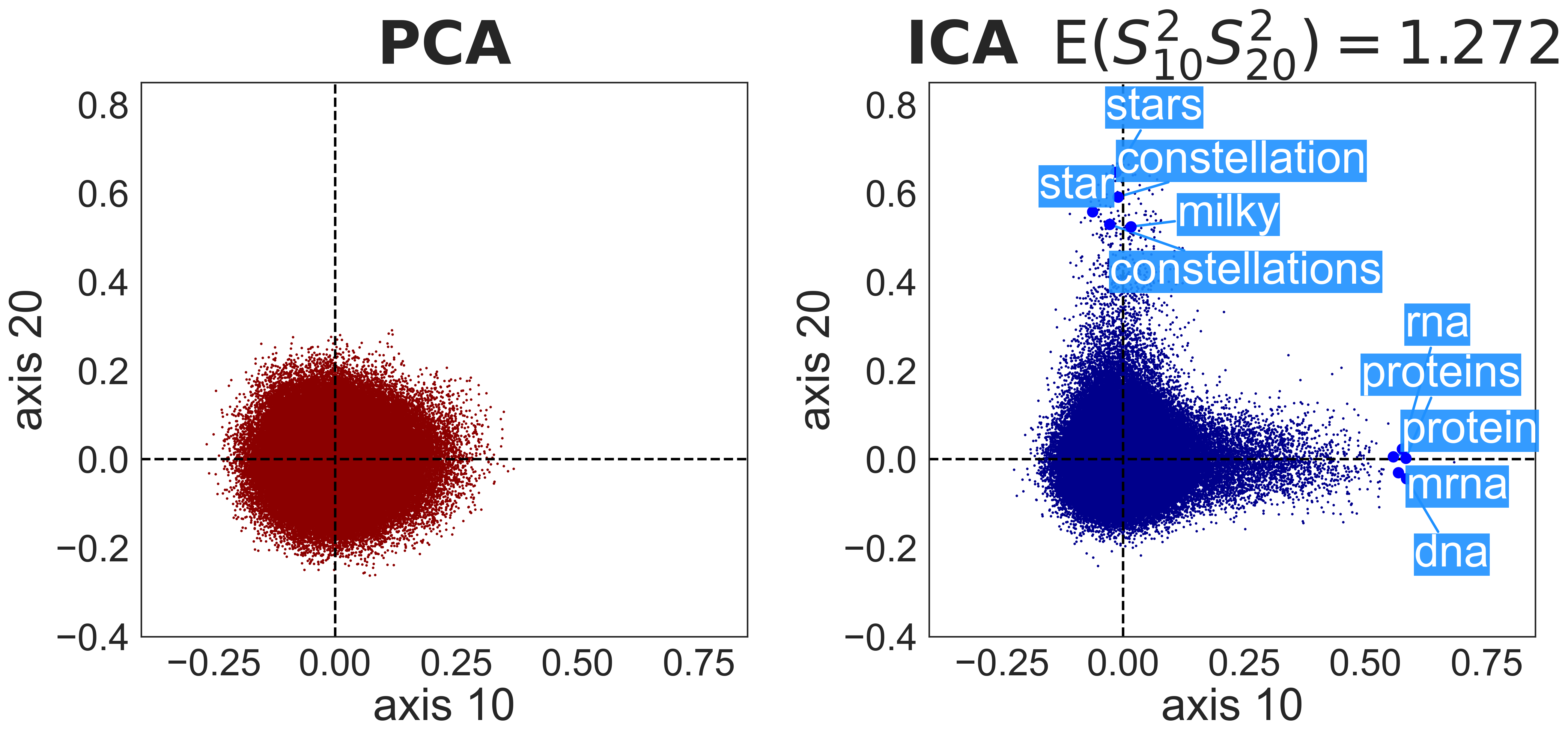}
    \caption{Scatterplots of normalized word embeddings along the 10th and 20th axes. The axes for PCA and ICA-transformed embeddings were arranged in descending order of variance and skewness, respectively. In both transformations, the components are uncorrelated.
    }
    \label{fig:scatter_pca-vs-ica}
\end{figure}

Figure~\ref{fig:scatter_pca-vs-ica} shows that ICA can find the `spiky and interpretable shape' of the embedding distribution (e.g., ``biology'' and ``stars'' for the 10th and 20th axes, respectively), but PCA cannot.
This is because ICA determines the coordinate axes toward high non-gaussianity, while PCA only considers variance information.

\section{Higher-Order Correlations Among Estimated Independent Components} \label{sec:dependency}

\paragraph{Non-Independence in Real-World Data.}
The `independent components' estimated by ICA on real-world data are uncorrelated but not completely independent, with dependencies existing between components~\cite{hyvarinen01topographic, sasaki13correlated, sasaki14estimating}.
This is because ICA assumes a linear decomposition into independent components, an assumption frequently violated in reality.

\paragraph{Higher-Order Correlation.}
To quantify non-independencies, methods like mutual information and Hilbert-Schmidt Independence Criterion (HSIC)~\cite{hsic} exist. 
Here we use the higher-order correlation, the simplest measure in terms of computation and formulation. This measure is expressed as follows:

\begin{equation} \label{eq:quadratic-interaction}
\mathrm{E}(S_{i}^2 S_{j}^2) = \frac{1}{n}\sum_{t=1}^{n} \mathbf{S}_{t,i}^{2} \mathbf{S}_{t,j}^{2}.
\end{equation}
Here, $\mathbf{S}$ is the whitened matrix\footnote{The components are (i)~centered:~$\mathrm{E}(S_{i})=0$, the mean of each component is  $0$,  (ii)~scaled:~$\mathrm{E}(S_{i}^2)=1$, the variance of each component is $1$, and (iii)~uncorrelated:~$\mathrm{E}(S_{i}S_{j})=0$, the correlations are all zero.}.
This can also be interpreted as the covariance between $S_{i}^2$ and $S_{j}^2$, plus one, as $\mathrm{cov}(S_{i}^2,S_{j}^2) = 
\mathrm{E}((S_{i}^2-1)(S_{j}^2-1)) = \mathrm{E}(S_{i}^2 S_{j}^2) - 1$.
If $S_{i}$ and $S_{j}$ are independent of each other, then $\mathrm{E}(S_{i}^2 S_{j}^2)= \mathrm{E}(S_{i}^2)\mathrm{E}(S_{j}^2) = 1$. 
Thus, the deviation of $\mathrm{E}(S_{i}^2 S_{j}^2)$ from $1$ is the degree of dependence between $S_{i}$ and $S_{j}$.

\begin{figure}[ht]
    \centering
    \includegraphics[width=0.92\columnwidth]{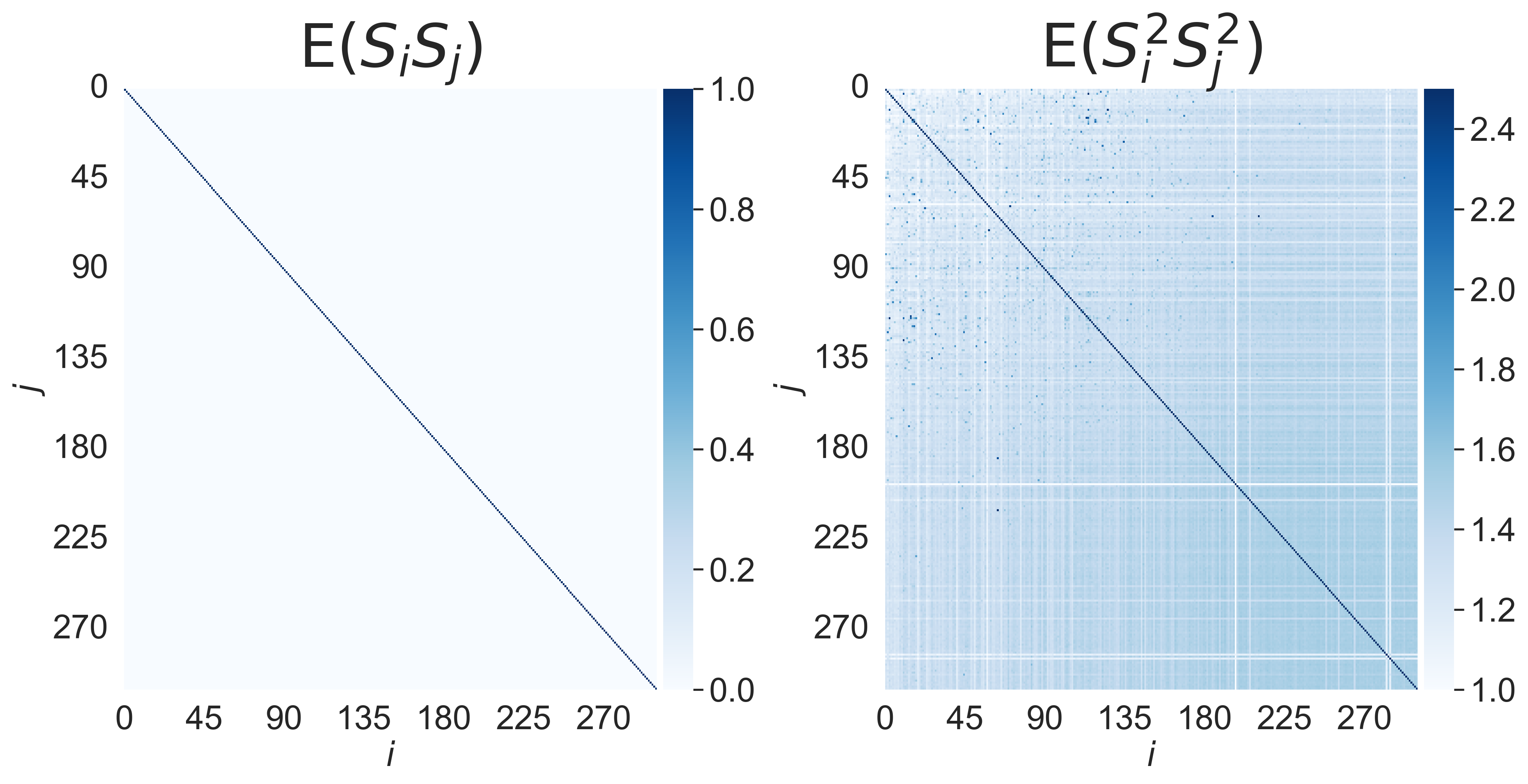}
    \caption{Heatmaps of the correlation coefficient $\mathrm{E}(S_{i}S_{j})$ and the higher-order correlation \Esisj{} of component pairs $(S_{i},S_{j})$ from ICA on 300-dimensional SGNS embeddings. See Appendix~\ref{app:higher-order-correlations} for details.}
    \label{fig:heatmap_3}
\end{figure}

Figure~\ref{fig:heatmap_3} shows that the estimated independent components of the embeddings are uncorrelated but not completely independent, with varying degrees of higher-order correlations across pairs.
These \Esisj{} values provide a useful metric of association, as demonstrated in the following section.

\begin{table*}[ht]
\centering
\begin{adjustbox}{width=\linewidth}
 \begin{tabular}{llllllllllllllllllll}
  \toprule
\multicolumn{2}{l}{$\mathrm{E}(S_{0}^2 S_{82}^2) = 1.927$} & \multicolumn{2}{l}{$\mathrm{E}(S_{6}^2 S_{96}^2) = 2.032$} & \multicolumn{2}{l}{$\mathrm{E}(S_{12}^2 S_{66}^2) = 1.975$} & \multicolumn{2}{l}{$\mathrm{E}(S_{16}^2 S_{118}^2) = 2.124$} & \multicolumn{2}{l}{$\mathrm{E}(S_{56}^2 S_{126}^2) = 1.861$} & \multicolumn{2}{l}{$\mathrm{E}(S_{63}^2 S_{210}^2) = 2.964$}\\
Axis 0 & Axis 82 & Axis 6 & Axis 96 & Axis 12 & Axis 66 & Axis 16 & Axis 118 & Axis 56 & Axis 126 & Axis 63 & Axis 210\\
\cmidrule(lr){1-2}\cmidrule(lr){3-4}\cmidrule(lr){5-6}\cmidrule(lr){7-8}\cmidrule(lr){9-10}\cmidrule(lr){11-12}
dishes & beer & el & o & rabbi & judah & blood & disorder & cpu & pointer & organization & unesco\\
sauce & beers & spanish & portuguese & talmud & israelites & organs & mental & microprocessor & return & international & itu\\
fried & ale & nacional & paulo & rabbis & yahweh & liver & disorders & processor & string & organizations & interpol\\
dish & brewing & jos & rio & torah & elisha & kidney & symptoms & cpus & pointers & interpol & observer\\
cooked & yeast & de & portugal & jewish & isaiah & tissue & bipolar & intel & node & standardization & temporary\\
\cmidrule(lr){1-2}\cmidrule(lr){3-4}\cmidrule(lr){5-6}\cmidrule(lr){7-8}\cmidrule(lr){9-10}\cmidrule(lr){11-12}
\multicolumn{2}{l}{$\mathrm{E}(S_{0}^2 S_{23}^2) = 0.990$} & \multicolumn{2}{l}{$\mathrm{E}(S_{6}^2 S_{13}^2) = 0.992$} & \multicolumn{2}{l}{$\mathrm{E}(S_{12}^2 S_{57}^2) = 0.993$} & \multicolumn{2}{l}{$\mathrm{E}(S_{16}^2 S_{57}^2) = 0.996$} & \multicolumn{2}{l}{$\mathrm{E}(S_{56}^2 S_{197}^2) = 0.982$} & \multicolumn{2}{l}{$\mathrm{E}(S_{63}^2 S_{18}^2) = 1.073$}\\
Axis 0 & Axis 23 & Axis 6 & Axis 13 & Axis 12 & Axis 57 & Axis 16 & Axis 57 & Axis 56 & Axis 197 & Axis 63 & Axis 18\\
\cmidrule(lr){1-2}\cmidrule(lr){3-4}\cmidrule(lr){5-6}\cmidrule(lr){7-8}\cmidrule(lr){9-10}\cmidrule(lr){11-12}
dishes & statesman & el & windows & rabbi & s & blood & s & cpu & population & organization & actress\\
sauce & astronomer & spanish & os & talmud & and & organs & and & microprocessor & median & international & footballer\\
fried & philosopher & nacional & unix & rabbis & was & liver & was & processor & estimated & organizations & musician\\
dish & johann & jos & linux & torah & in & kidney & in & cpus & residing & interpol & actor\\
cooked & mathematician & de & microsoft & jewish & by & tissue & by & intel & total & standardization & singer\\
  \bottomrule
 \end{tabular}
 \end{adjustbox}
  \caption{
  (Top Row) Six randomly selected pairs of components from the top 50 pairs with the highest $|\mathrm{E}(S_i^2 S_j^2) - 1|$ values. For each component, the top 5 words (frequency $n_w \geq 100$ in text8) with the largest component values are listed. (Bottom Row) Component pairs with small $\mathrm{E}(S_{i}^2 S_{k}^2)$ values. For each component $S_i$ with the smaller axis number in the pairs \sisj{} in the top row, a component $S_k$ with the smallest value of $|\mathrm{E}(S_i^2 S_k^2) - 1|$ was selected.
  }
\label{tab:example_energy}
\end{table*}

\begin{table}[ht]
    \centering
    \begin{adjustbox}{width=\linewidth}
    \begin{tabular}{lccccc}
    \toprule
                     & $k=1$ & $k=2$ & $k=3$ & $k=4$ & $k=5$\\
        \midrule
        List-2 (top-$k$)       & \textbf{69.0} & \textbf{65.0} & \textbf{64.0} & \textbf{64.5} & \textbf{56.5}\\
        List-3 (bottom 30\%)  & 27.0 & 33.0 & 32.5 & 33.5 & 40.5\\
        Can't decide & 4.0 & 2.0 & 3.5 & 2.0 & 3.0 \\
    \bottomrule
    \end{tabular}
    \end{adjustbox}
    \caption{The percentage of each list judged by the GPT model to be more semantically related to List-1.}
    \label{tab:gpt_eval_ica}
\end{table}

\begin{table*}[ht]
\centering
\begin{adjustbox}{width=\linewidth}
 \begin{tabular}{lrlrlrlrlrlrlrlrlrlrlrlr}
  \toprule
\multicolumn{2}{l}{$\mathrm{E}(S_{10}^2 S_{2}^2) = 2.323$} & \multicolumn{2}{l}{$\mathrm{E}(S_{10}^2 S_{16}^2) = 1.947$} & \multicolumn{2}{l}{$\mathrm{E}(S_{10}^2 S_{160}^2) = 1.811$} & \multicolumn{2}{l}{$\mathrm{E}(S_{27}^2 S_{11}^2) = 1.643$} & \multicolumn{2}{l}{$\mathrm{E}(S_{27}^2 S_{64}^2) = 1.997$} & \multicolumn{2}{l}{$\mathrm{E}(S_{27}^2 S_{104}^2) = 1.605$}\\
Axis $10$ & \ml{Axis $2$} & Axis $10$ & \ml{Axis $16$} & Axis $10$ & \ml{Axis $160$} & Axis $27$ & \ml{Axis $11$} & Axis $27$ & \ml{Axis $64$} & Axis $27$ & \ml{Axis $104$}\\
\cmidrule(lr){1-1}\cmidrule(lr){2-2}\cmidrule(lr){3-3}\cmidrule(lr){4-4}\cmidrule(lr){5-5}\cmidrule(lr){6-6}\cmidrule(lr){7-7}\cmidrule(lr){8-8}\cmidrule(lr){9-9}\cmidrule(lr){10-10}\cmidrule(lr){11-11}\cmidrule(lr){12-12}
dna & \ml{acid} & dna & \ml{blood} & dna & \ml{evolution} & greek & \ml{gaius} & greek & \ml{goddess} & greek & \ml{archaeological}\\
proteins & \ml{hydrogen} & proteins & \ml{organs} & proteins & \ml{evolutionary} & greece & \ml{caesar} & greece & \ml{gods} & greece & \ml{neolithic}\\
rna & \ml{acids} & rna & \ml{liver} & rna & \ml{darwin} & athens & \ml{augustus} & athens & \ml{deity} & athens & \ml{bc}\\
mrna & \ml{oh} & mrna & \ml{kidney} & mrna & \ml{selection} & athenian & \ml{lucius} & athenian & \ml{deities} & athenian & \ml{pottery}\\
\cmidrule(lr){1-2}\cmidrule(lr){3-4}\cmidrule(lr){5-6}\cmidrule(lr){7-8}\cmidrule(lr){9-10}\cmidrule(lr){11-12}
$w_k$ & $\mathbf{S}_{k,10}^{2}\mathbf{S}_{k,{2}}^{2}$ & $w_k$ & $\mathbf{S}_{k,10}^{2}\mathbf{S}_{k,{16}}^{2}$ & $w_k$ & $\mathbf{S}_{k,10}^{2}\mathbf{S}_{k,{160}}^{2}$ & $w_k$ & $\mathbf{S}_{k,27}^{2}\mathbf{S}_{k,{11}}^{2}$ & $w_k$ & $\mathbf{S}_{k,27}^{2}\mathbf{S}_{k,{64}}^{2}$ & $w_k$ & $\mathbf{S}_{k,27}^{2}\mathbf{S}_{k,{104}}^{2}$\\
\cmidrule(lr){1-2}\cmidrule(lr){3-4}\cmidrule(lr){5-6}\cmidrule(lr){7-8}\cmidrule(lr){9-10}\cmidrule(lr){11-12}
ribose & 3755.7 & adenylate & 2079.8 & utr & 2381.5 & laertius & 898.3 & demeter & 2348.5 & tiryns & 1690.6\\
deoxyribose & 2963.9 & effectors & 1842.5 & reticulum & 1942.0 & preveza & 788.0 & hephaestus & 2204.3 & knossos & 1348.1\\
phosphodiester & 2850.2 & antisense & 1639.9 & genomic & 1668.6 & xanthippus & 764.2 & hestia & 2021.5 & mycenaean & 1205.6\\
biosynthesis & 2510.1 & cyclase & 1638.9 & homozygous & 1599.1 & rhadamanthus & 735.5 & hera & 1744.6 & lendering & 1124.7\\
methyltransferase & 2482.9 & myosin & 1201.8 & cleaved & 1181.0 & thracians & 711.8 & cronos & 1720.2 & hissarlik & 1103.1\\
pyrimidine & 2399.6 & axons & 1144.2 & tubulin & 1152.4 & alexandri & 705.2 & aphrodite & 1675.9 & melos & 1006.3\\
  \bottomrule
 \end{tabular}
 \end{adjustbox}
 \caption{For 6 component pairs \sisj{} selected from adjacent component pairs in the MST defined in Sec.~\ref{sec:visualization}, the top 6 words and their corresponding $\mathbf{S}_{t,i}^{2} \mathbf{S}_{t,j}^{2}$ values that contribute the most to the \Esisj{} value are shown.}
\label{tab:xy_words}
\end{table*}

\section{Interpretation of Higher-Order Correlations as Semantic Relevance} \label{sec:semantic_relevance}

\subsection{Degree of Semantic Relevance} \label{subsec:pair_example}
We show that the values of higher-order correlations \Esisj{} can be interpreted as the degree of associations between semantic components.
\paragraph{Results: Top Row of Table~\ref{tab:example_energy}.}
The meanings of each component, represented by the listed words in component pairs with high \Esisj{} values, are strongly related.
For example, focusing on Axis $0$ and Axis $82$, a pair with particularly large values of \Esisj{}, we can interpret that Axis $0$ has a meaning associated with ``dishes'' and Axis $82$ with ``beer'', suggesting that there is a semantic relationship between them.
\paragraph{Results: Bottom Row of Table~\ref{tab:example_energy}.}
On the other hand, for component pairs with \Esisj{} values close to $1$, indicating that the components are considered independent, there is no clear relevance between the meanings of the components.
For example, looking at the pair of Axis $0$ and Axis $23$, which has a small \Esisj{} value, we can interpret that Axis $0$ represents ``dishes'' and Axis $23$ represents ``polymath'', and there is no direct semantic relationship between them. 

Detailed results are shown in Appendix~\ref{appendix:experiment_results}.

\subsection{Quantitative Evaluation via GPT-4o mini}
We conducted experiments to quantitatively evaluate whether higher-order correlations between ICA components represent semantic relationships.

\paragraph{Settings.}
Our experimental procedure was as follows. We first selected the top 100 ICA components, ranked by skewness. For each component $i$ ($i = 0, \ldots, 99$), we created three word lists: Word list-1 comprised the top 5 words from component $i$, Word list-2 contained the top 5 words from the $k$-th most correlated component with component $i$ ($k = 1, \ldots, 5$), and Word list-3 consisted of the top 5 words from a randomly selected low-correlation component (chosen from the bottom 30\% of correlated components).
Using these lists, we generated pairs (list-1, list-2) and (list-1, list-3), and queried GPT-4o mini to determine which pair was more semantically related\footnote{To mitigate potential biases, we randomly shuffled the order of the lists in the pairs and repeated this process with the reversed order: (list-1, list-3) and (list-1, list-2).}.
This procedure was executed for all 100 components, resulting in 200 total comparisons for each value of $k$ from 1 to 5.
The specific prompt used for GPT-4o mini model is provided in Appendix~\ref{appendix:gpt-evaluation}.

\paragraph{Results and Discussion.}
Table~\ref{tab:gpt_eval_ica} shows the result of the experiment.
We can see that component pairs with higher-order correlations tend to be more semantically related (69.0\% for $k=1$ vs 27.0\% for bottom 30\%), and that semantic relatedness gradually declines as correlation decreases (69.0\% at $k=1$ to 56.5\% at $k=5$).
These results quantitatively demonstrate that higher-order correlations between ICA components effectively reflect semantic relatedness between corresponding words.

\subsection{Decomposition of Semantic Relevance} \label{subsec:words_example}

\begin{figure}[!t]
    \centering
    \includegraphics[width=0.90\columnwidth]{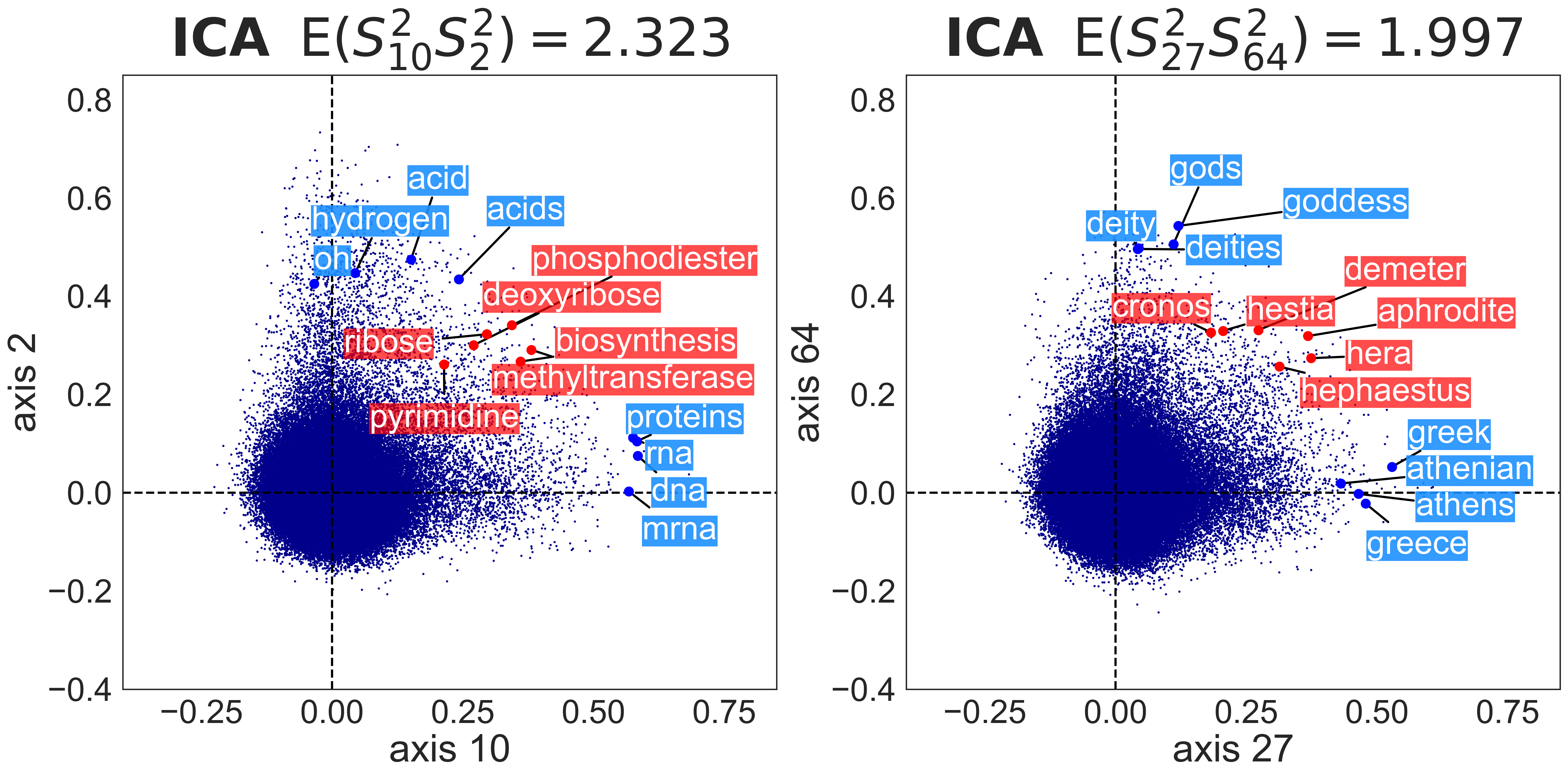}
    \caption{Scatter plots of normalized word embeddings for axis pairs (10, 2) and (27, 64) with large values of higher-order correlations. Blue-labeled words are the top 4 words for each axis's component values, while red-labeled words are the top 6 words for the values of $\mathbf{S}_{t,i}^{2} \mathbf{S}_{t,j}^{2}$. See Appendix~\ref{app:higher-order-correlations} for all the pairs in Table~\ref{tab:xy_words}.}
    \label{fig:table3_maintext}
\end{figure}

For a component pair \sisj{}, words $w_t$ with large values of $\mathbf{S}_{t,i}^{2} \mathbf{S}_{t,j}^{2}$ are considered to make a significant contribution to the higher-order correlation $\mathrm{E}(S_i^2 S_j^2)=\frac{1}{n}\sum_{t=1}^{n} \mathbf{S}_{t,i}^{2} \mathbf{S}_{t,j}^{2}$. 
Here we investigate words that significantly contribute to the \Esisj{} values and gain a more concrete understanding of the relationships between components.

\paragraph{Results.}
Table~\ref{tab:xy_words} presents component pairs selected from the maximum spanning tree $T_{150}$ (Sec.~\ref{sec:visualization}) and words significantly contributing to their \Esisj{} values.
These words often relate to the meanings of both components, demonstrating additive compositionality.
For example, in the Axis 10 and Axis 2 pair, words like \textit{ribose}, \textit{deoxyribose}, \textit{phosphodiester}, \textit{biosynthesis}, \textit{methyltransferase}, and \textit{pyrimidine} notably contribute to the \Esisj{} value, linking biomolecules and chemical components.
Detailed results are shown in Appendix~\ref{appendix:experiment_results}.

\paragraph{Visualization.}
Figure~\ref{fig:table3_maintext} shows word embedding scatter plots for axis pairs (10, 2) and (27, 64) with large higher-order correlations to illustrate the distribution of words with significant contributions to higher-order correlations.
Unlike a typical independent component scatter plot (Fig.~\ref{fig:scatter_pca-vs-ica}), these exhibit many words with large component values in both axes, reflecting the meanings of both axes and demonstrating the additive compositionality of embeddings.
For the (10, 2) pair, words that notably contribute to the \Esisj{} (\textit{ribose}, \textit{deoxyribose}, \textit{phosphodiester}, \textit{biosynthesis}, \textit{methyltransferase}, and \textit{pyrimidine}) appear with significant values in both components.
This abundance of words sharing both semantic components is characteristic of pairs with large higher-order correlations.
Detailed results are shown in Appendix~\ref{app:higher-order-correlations}.

\section{Visualization of Non-Independence Structure} \label{sec:visualization}

\begin{figure*}[ht]
    \centering
    \includegraphics[width=0.90\linewidth]{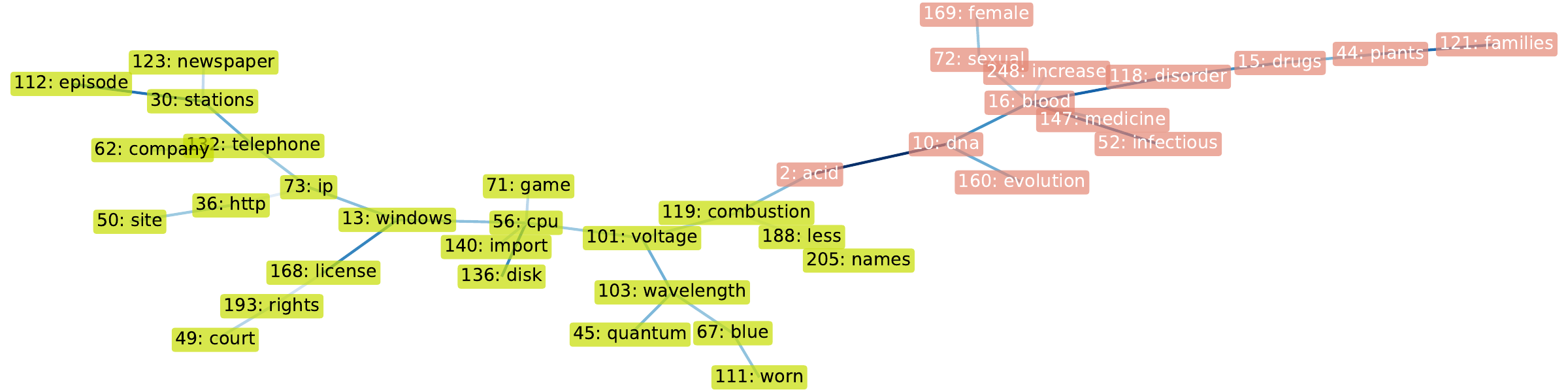}
    \includegraphics[width=0.90\linewidth]{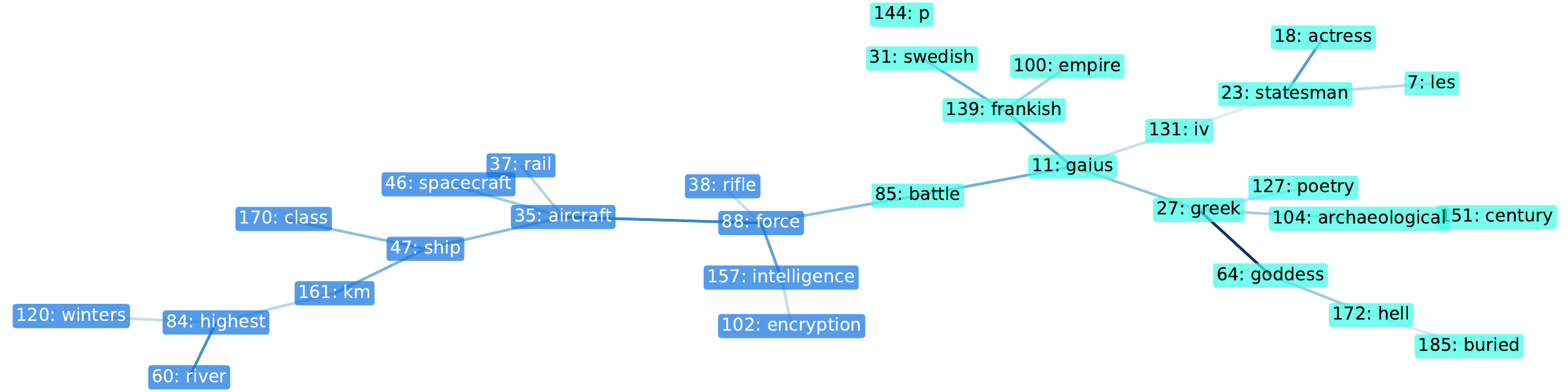}
    \caption{Subtrees of MST $T_{150}$ defined in Sec.~\ref{sec:visualization}. Each node represents an independent component $S_k$ (i.e., Axis $k$) estimated by ICA. The label of each node is ``$k$ : $\mathrm{TopWord}(k)$'', where $\mathrm{TopWord}(k)$ is the word with the largest component value along axis $k$ among words with frequency $n_w \geq 100$ in the text8 corpus. The color of the edge between nodes $(i, j)$ represents the magnitude of the \Esisj{} value between the components, with darker edge colors indicating larger values.}
    \label{fig:mst}
\end{figure*}

In this section, we construct a maximum spanning tree (MST) based on higher-order correlations to visualize the non-independence between estimated independent components.

\paragraph{Settings.}
The 300 ICA components, originally ordered by skewness with $i = 0, \cdots, 299$, were re-sorted in descending order of semantic component consistency scores to prioritize axes that are more easily interpretable as specific semantic components.
The semantic component consistency scores were determined by a word intrusion task~\cite{chang-2009-reading}. A higher consistency score indicates easier interpretability. Details of the scoring methods are provided in Appendix~\ref{appendix:intrusion}. We introduce the notation $\sigma$ to map the order of consistency scores to the original axis numbers in the skewness order:
$\sigma(j)$ represents the axis number in the skewness sort for the axis with the $j$-th highest consistency score.
Then, we consider a weighted complete graph $G_{150}$, with 150 components having high consistency scores $S_{\sigma(i)}~(i \in {0, \cdots, 149})$ as nodes.
For the edge between the node pair $(S_i, S_j)$, we set $c_{ij}=\mathrm{E}(S_i^2 S_j^2)$ as the weight.
To visualize and interpret $G_{150}$, we compute the maximum spanning tree (MST)\footnote{
We used \texttt{minimum\_spanning\_tree} implemented in \texttt{NetworkX}~\cite{networkx} for the computation of the MST. See Appendix~\ref{appendix:mst} for details.
}
$T_{150}$, a spanning tree that maximizes the sum of $c_{ij}$ in the graph $G_{150}$.
MST was relatively more interpretable than other subgraphs of graph $G_{150}$, providing a good balance between visibility and element relationships. 

\paragraph{Interpretation of the MST.}
The MST $T_{150}$ represents a graph structure expressing the non-independence between estimated independent components.
Since the edges in $T_{150}$ connect component pairs with large higher-order correlations, we can interpret that there is a strong relationship between the components connected by these edges.
Furthermore, the subtrees of $T_{150}$ represent groups of semantically related components, and the components within these groups tend to have similar meanings.

\paragraph{Results and Discussion.}
Figure~\ref{fig:mst} shows a part of the MST $T_{150}$; the entire MST $T_{150}$ is exhibited in Appendix~\ref{appendix:mst}.
The colors correspond to the clusters obtained by applying spectral clustering\footnote{We used \texttt{SpectralClustering} implemented in \texttt{scikit-learn}~\cite{sklearn}.}~\cite{spectral-clustering} to $T_{150}$.
The weights used for clustering are the higher-order correlations.
From the MST, we can infer structures such as connections and groupings of meanings among three or more components\footnote{Furthermore, in Appendix~\ref{appendix:dimension-compression}, we conducted a dimensionality reduction experiment that numerically demonstrates how the MST effectively represents a significant structure among the components.}.
For example, semantically related component pairs such as (\textit{2: dna, 10: acid}) in the pink cluster and (\textit{27: greek, 64: goddess}) in the cyan cluster are connected by edges in $T_{150}$.
Additionally, groups such as $\{$\textit{168: license, 13: windows, 56: cpu}$\}$ in the yellow cluster and $\{$\textit{46: spacecraft, 35: aircraft, 47: ship}$\}$ in the blue cluster form semantic clusters as sets of nodes connected by edges. The components within these groups can be interpreted as having meanings related to ``computer'' and ``vehicle'', respectively.

\section{Conclusion}
Both ICA and PCA transformations make the components uncorrelated. ICA goes further by making the components nearly independent, but some non-independence still remains.
In this study, we used higher-order correlations to quantify the non-independence between the components in the ICA-transformed word embeddings.
By interpreting these as the semantic associations between the components and visualizing the overall structure, we can gain a deeper understanding of the latent semantic structure within the embeddings.

\clearpage

\section*{Limitations}
\begin{itemize}
\item The embeddings used in the experiments are limited to SGNS word embeddings. For a more thorough analysis, it is necessary to conduct experiments using various types of embeddings.
\item For large embedding matrices with a high number of data points $n$, ICA may fail to converge within a practical timeframe. To overcome this, we suggest using subsampled data to estimate the ICA transformation matrix, which can then be applied to unseen embedding vectors.
\item When $n$ is large, calculating higher-order correlations (eq. \ref{eq:quadratic-interaction}) may become computationally intensive. This calculation is similar to the computation of the variance-covariance matrix and can be approximated by subsampling data points. Further speedup can be achieved by parallelization of the computation.
\item Since ICA leverages the non-Gaussianity of embedding distributions, it is not suitable for analysis if the original embeddings follow a multivariate Gaussian distribution.
\end{itemize}

\section*{Ethics Statement}
This study complies with the \href{https://www.aclweb.org/portal/content/acl-code-ethics}{ACL Ethics Policy}.

\section*{Acknowledgements}
This study was partially supported by JSPS KAKENHI 22H05106, 23H03355, JST CREST JPMJCR21N3, JST BOOST JPMJBS2407, JST SPRING JPMJSP2110.

\section*{Code Availability}
Code is available at \url{https://github.com/momoseoyama/hoc}.

\bibliography{anthology, custom}
\bibstyle{acl_natbib}

\appendix

\section{Details of Experimental Settings} \label{appendix:settings}

The word embeddings used in the experiments were trained using Skip-gram with Negative Sampling (SGNS).
The parameters used to train SGNS are summarized in Table~\ref{tab:sgns-parameters}.
The corpus used for training is the text8 corpus~\cite{mahoney2011}, and the number of vocabulary words is $n=253{,}854$.

\begin{table}[ht]
\small
\centering
    \begin{tabular}{lc}
        \toprule
        Dimensionality & 300 \\
        Epochs & 100 \\
        Window size $h$ & 10 \\
        Negative samples $\nu$ & 5 \\
        Learning rate & 0.025 \\
        Min count & 1\\
        \bottomrule
    \end{tabular}
\caption{SGNS parameters. }
\label{tab:sgns-parameters}
\end{table}

\section{Remarks on Axis $57$ in Figure~\ref{fig:heatmap_ica-vs-pca}} \label{app:axis57}

\begin{figure}[ht]
    \centering
    \includegraphics[width=0.95\columnwidth]{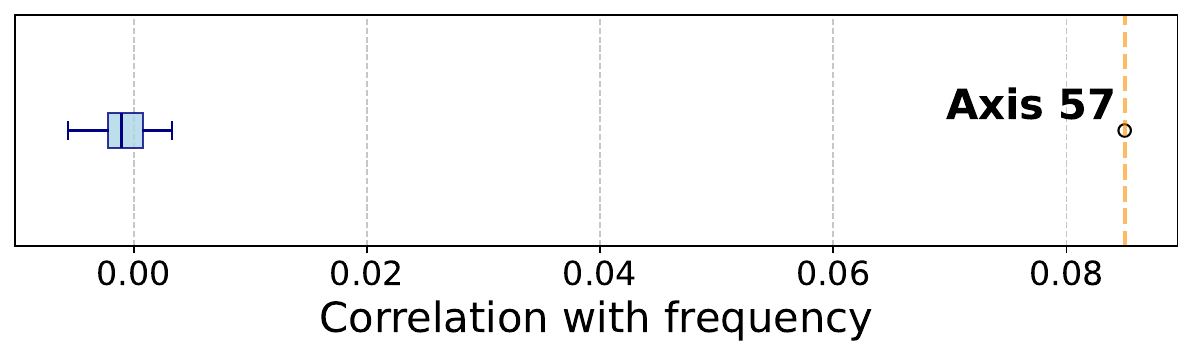}
    \caption{Boxplot of correlation coefficients between word frequency $n_w$ and the component values for the $0$th to $99$th axes of the ICA-transformed embeddings. Axis $57$ shows a particularly high correlation coefficient.}
    \label{fig:boxplot_correlation}
\end{figure}

\begin{figure}[t!]
    \centering
    \includegraphics[width=0.95\columnwidth]{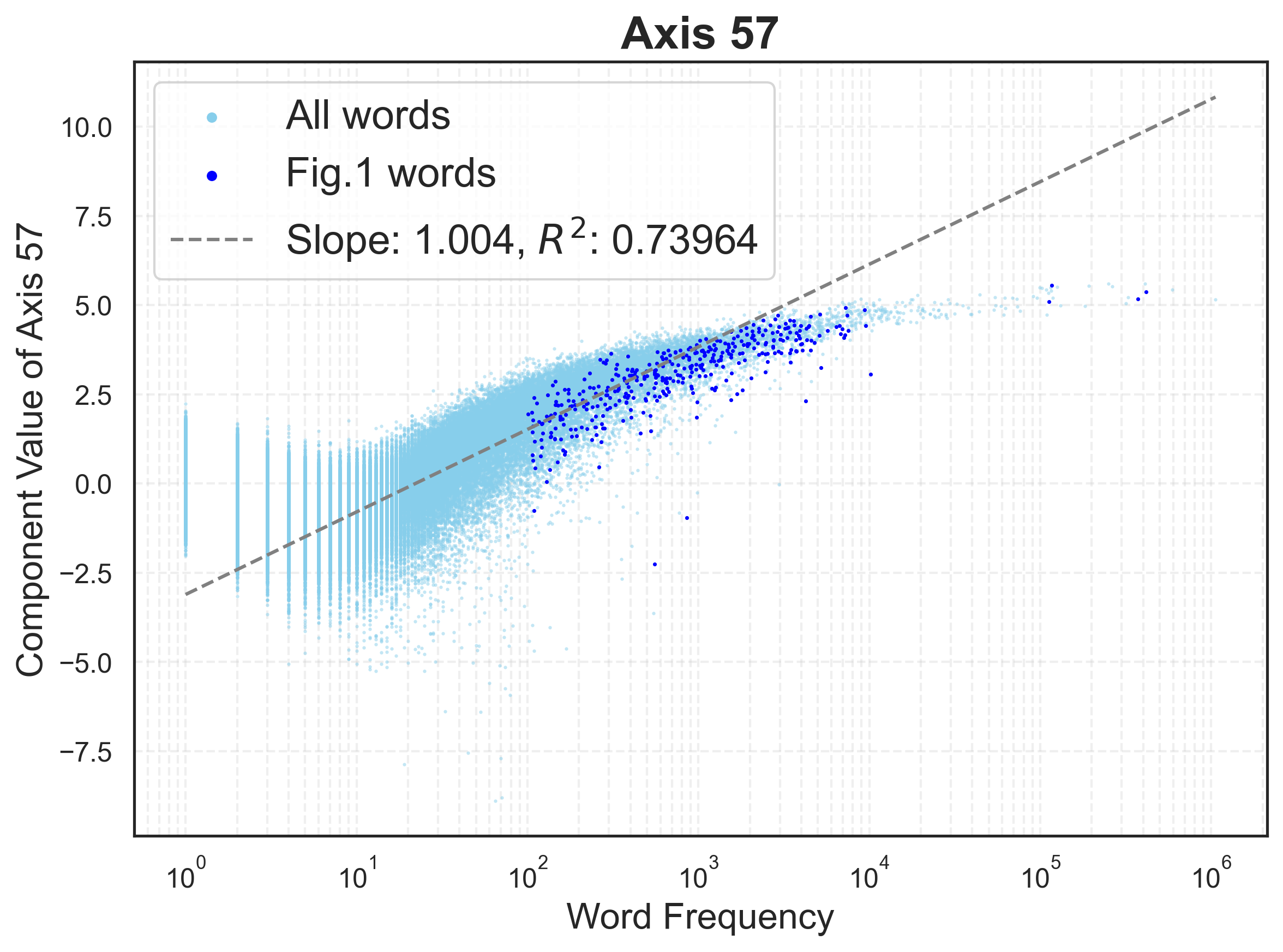}
    \caption{Scatter plot of word frequency $n_w$ versus the component values of the $57$th axis of the ICA-transformed embeddings. Words used in Fig.~\ref{fig:heatmap_ica-vs-pca} are highlighted in dark blue. The regression line and coefficient of determination were calculated for words with a frequency of $n_w \geq 10$.}
    \label{fig:scatter_axis57}
\end{figure}

\begin{figure}[t!]
    \centering
    \includegraphics[width=0.95\columnwidth]{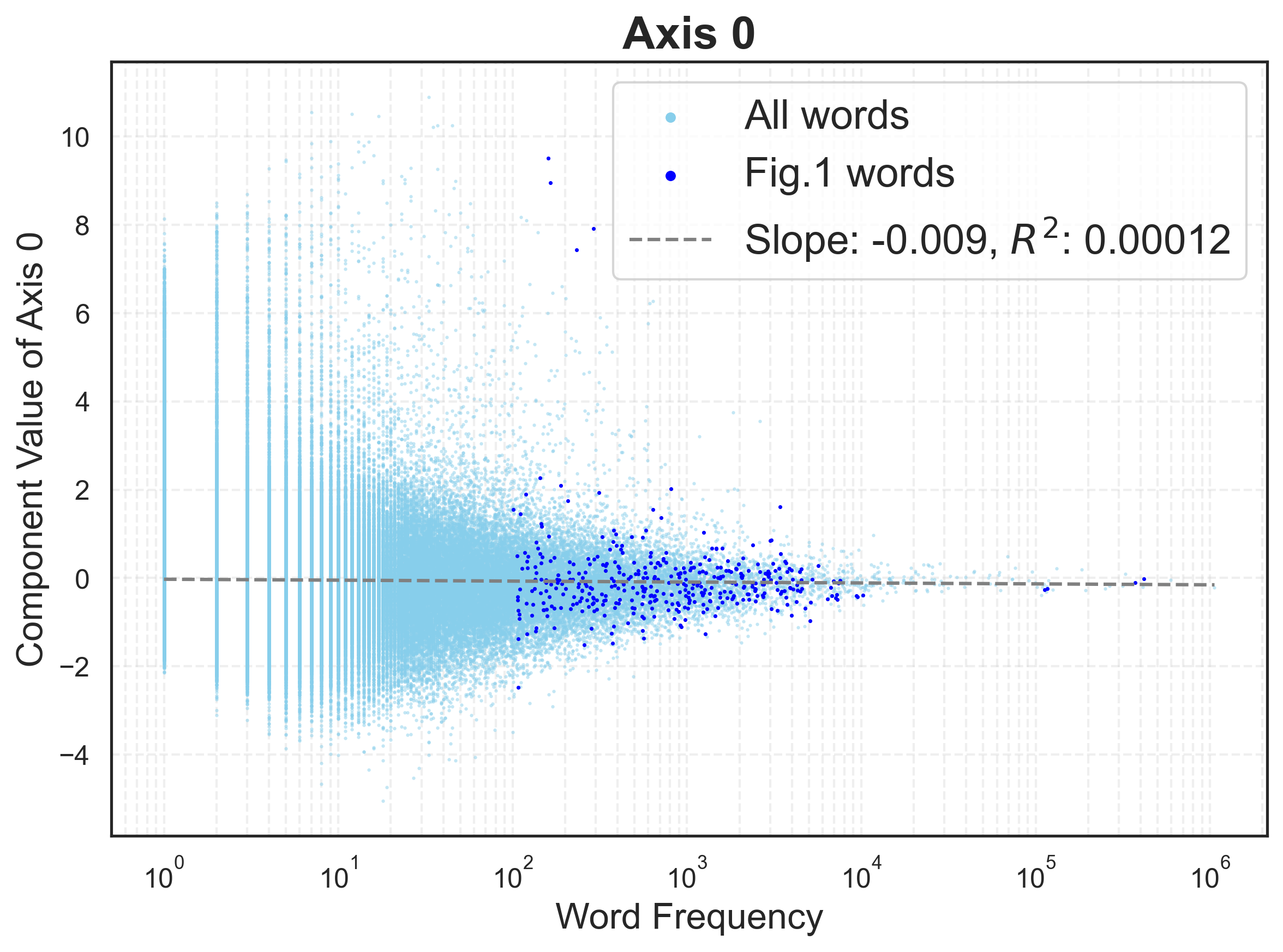}
    \caption{Scatter plot of word frequency $n_w$ versus the component values of the $0$th axis of the ICA-transformed embeddings. 
    The settings are the same as in Fig.~\ref{fig:scatter_axis57}.
    }
    \label{fig:scatter_axis0}
\end{figure}

An interesting vertical streak is observed in Axis $57$ of the heatmap for ICA-transformed embeddings in Fig.~\ref{fig:heatmap_ica-vs-pca}. This streak can be explained by several factors.
As shown in Fig.~\ref{fig:boxplot_correlation}, Axis $57$ exhibits a strong correlation between component values and word frequencies $n_w$, suggesting that Axis 57 is more associated with word frequency than with a specific semantic meaning.
Additionally, the words used in Fig.~\ref{fig:heatmap_ica-vs-pca} were selected from those appearing more than 100 times in the text8 corpus, resulting in a bias towards high-frequency words.
This tendency is further illustrated in Fig.~\ref{fig:scatter_axis57}, which demonstrates that words used in the heatmap $(n_w \geq 100)$ tend to have larger component values along Axis $57$.
In contrast, Fig.~\ref{fig:scatter_axis0} shows that for axes with weak correlation to word frequency, the words used in the heatmap do not exhibit notably large component values.
Consequently, large component values were observed along Axis 57 in the heatmap, a pattern that was unique to Axis 57 and not observed in other axes.

\begin{figure*}[!t]
    \centering
    \begin{subfigure}[b]{0.48\textwidth}
        \centering
        \includegraphics[width=0.95\textwidth]{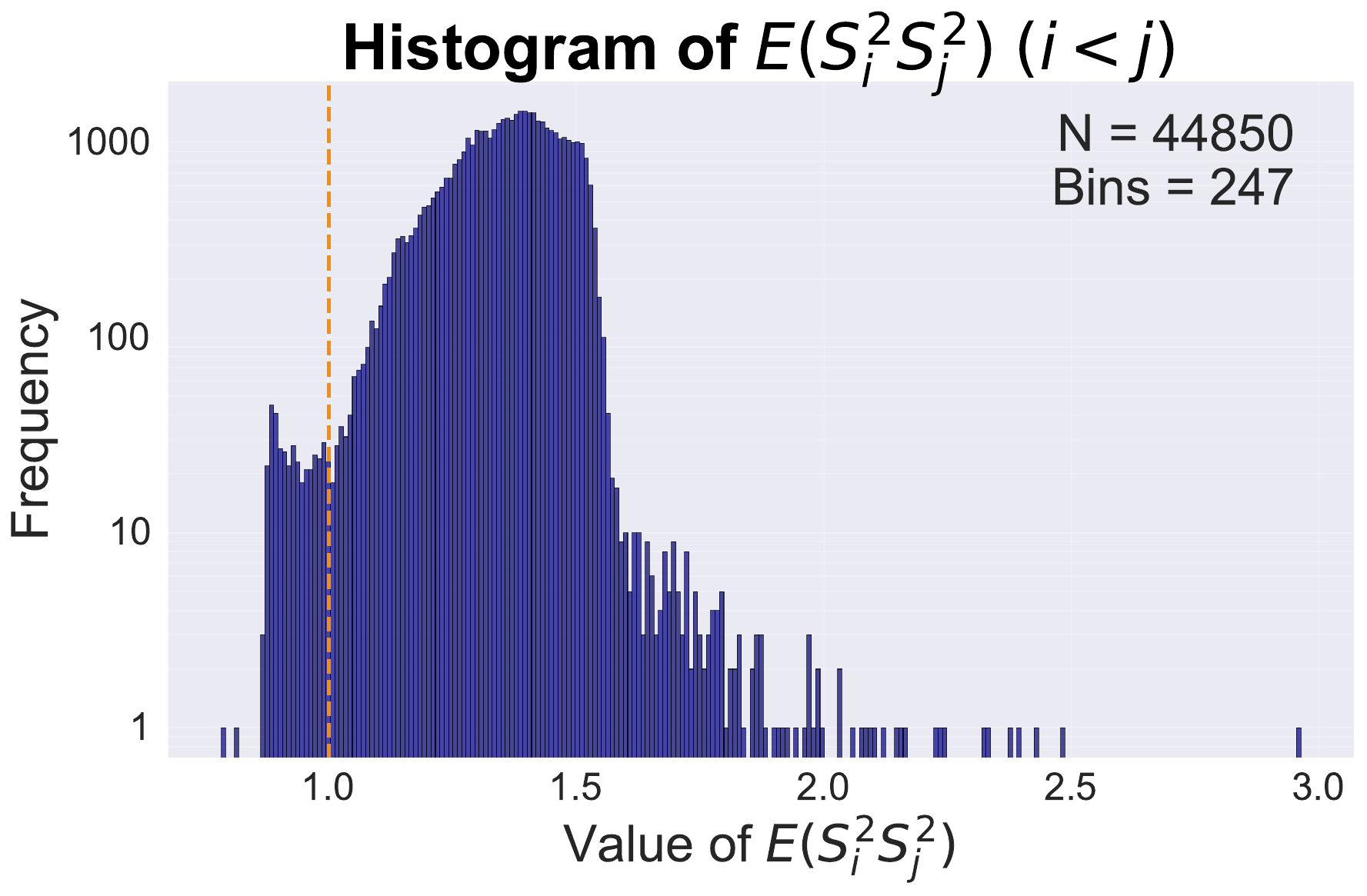}
        \caption{For $(i, j)$ pairs where $i < j$.}
        \label{fig:histogram_hoc_uptri}
    \end{subfigure}
    \hfill
    \begin{subfigure}[b]{0.48\textwidth}
        \centering
        \includegraphics[width=0.95\textwidth]{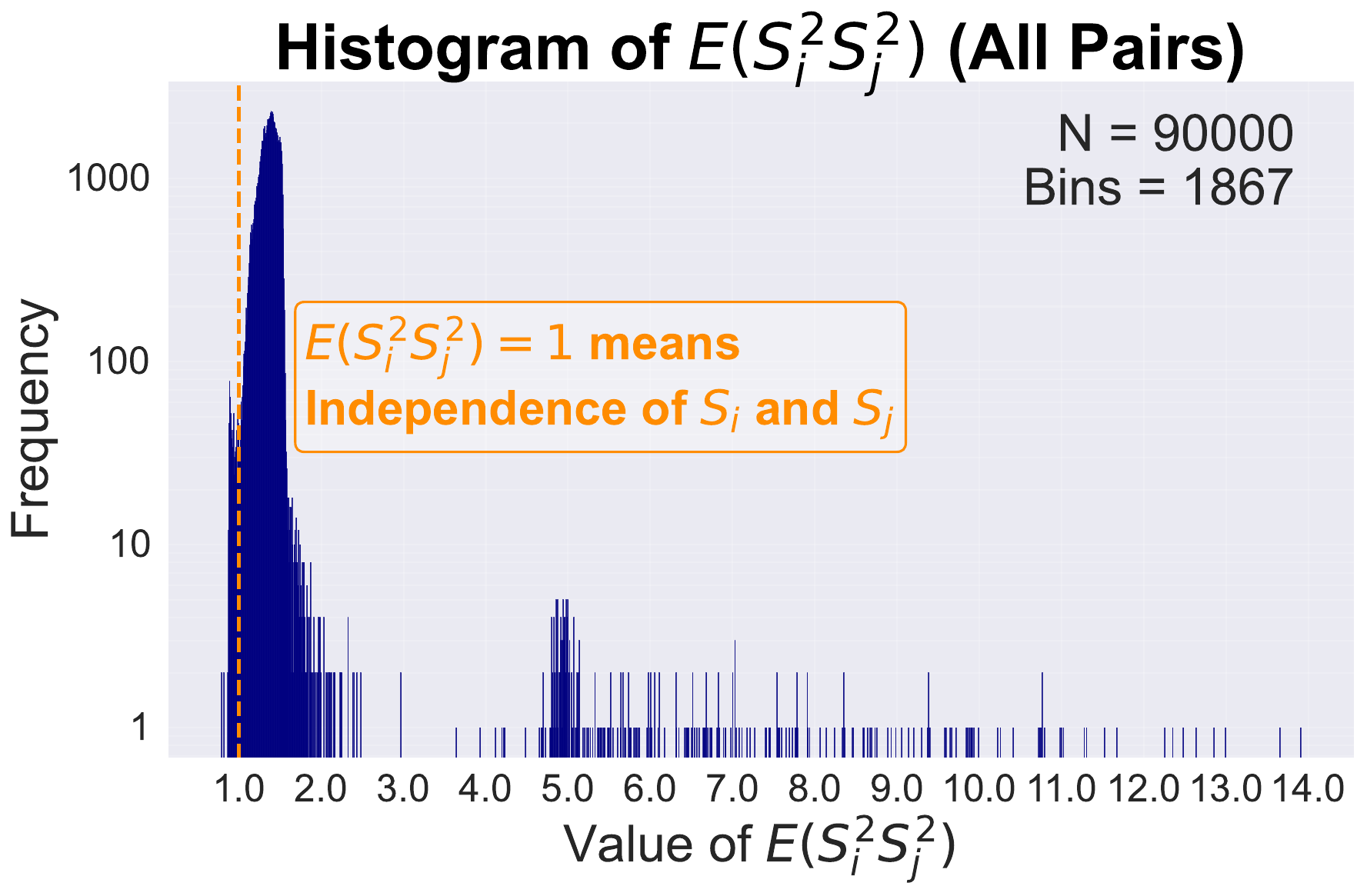}
        \caption{For all $(i, j)$ pairs, including $i=j$.}
        \label{fig:histogram_hoc_allpair}
    \end{subfigure}
    \caption{Histograms of higher-order correlations.}
    \label{fig:combined_histograms}
\end{figure*}

\section{Higher-Order Correlations} \label{app:higher-order-correlations}

\subsection{Distribution of Higher-Order Correlations} \label{app:histogram_hoc}
Figures~\ref{fig:histogram_hoc_uptri} and \ref{fig:histogram_hoc_allpair} show histograms of higher-order correlations $\mathrm{E}(S_i^2 S_j^2)$ for pairs where $i < j$ and for all pairs including $\mathrm{E}(S_i^4)$ where $i=j$, respectively.
While there are component pairs where $\mathrm{E}(S_i^2 S_j^2) < 1$, in Fig.~\ref{fig:heatmap_3}, the range of values was truncated between $1.0$ and $2.5$ for visualization purposes.

\subsection{Scatterplots for Independent Axes} \label{app:scatters_hoc}

\begin{figure}[!t]
    \centering
    \includegraphics[width=\columnwidth]{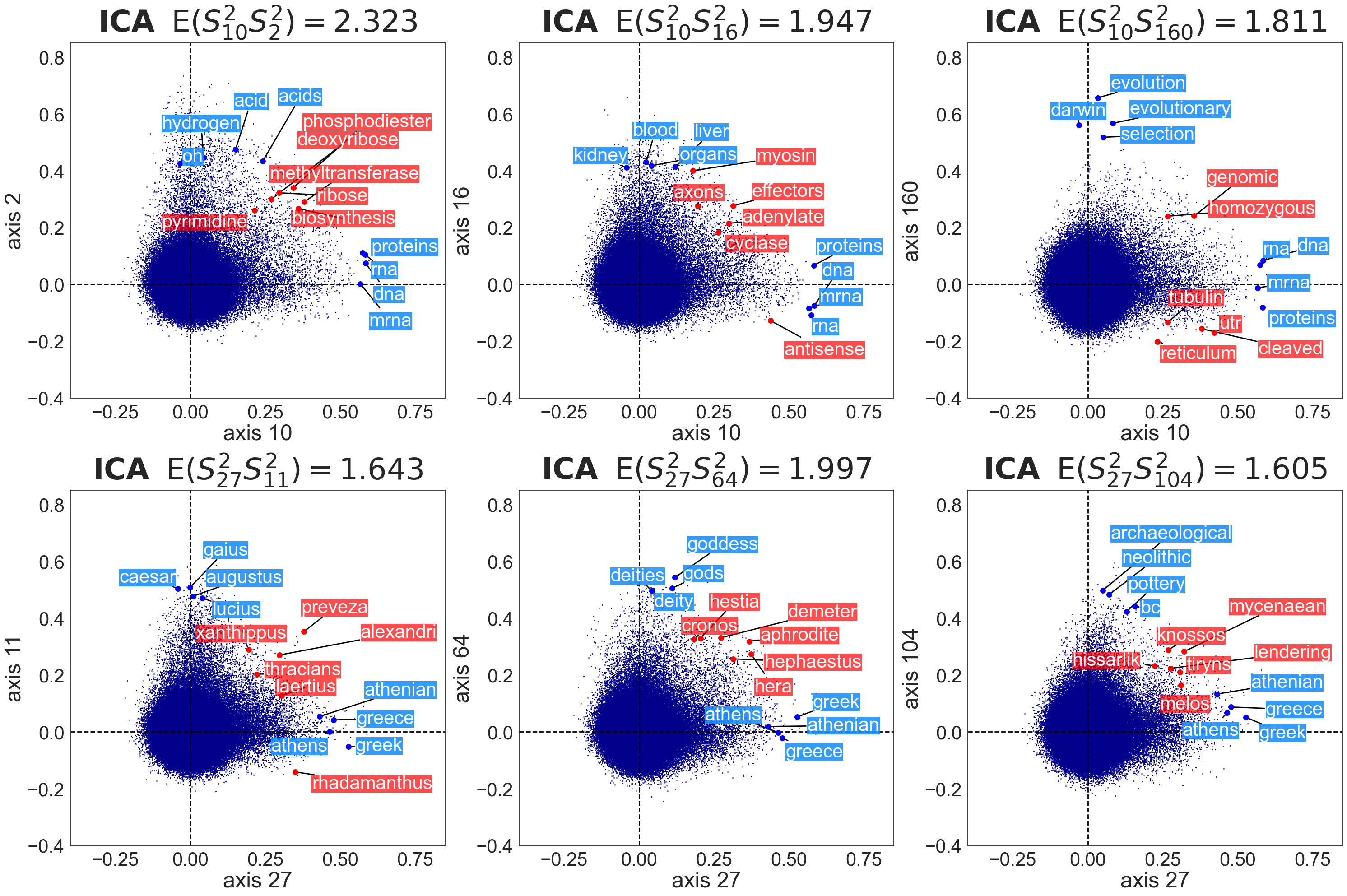}
     \caption{Scatter plots of normalized word embeddings for axis pairs in Table~\ref{tab:xy_words}. Blue-labeled words are the top 4 words for each axis's component values, while red-labeled words are the top 6 words for values of $\mathbf{S}_{t,i}^{2} \mathbf{S}_{t,j}^{2}$.}
    \label{fig:table3_appendix}
\end{figure}

\begin{figure*}[!t]
    \centering
    \includegraphics[width=\linewidth]{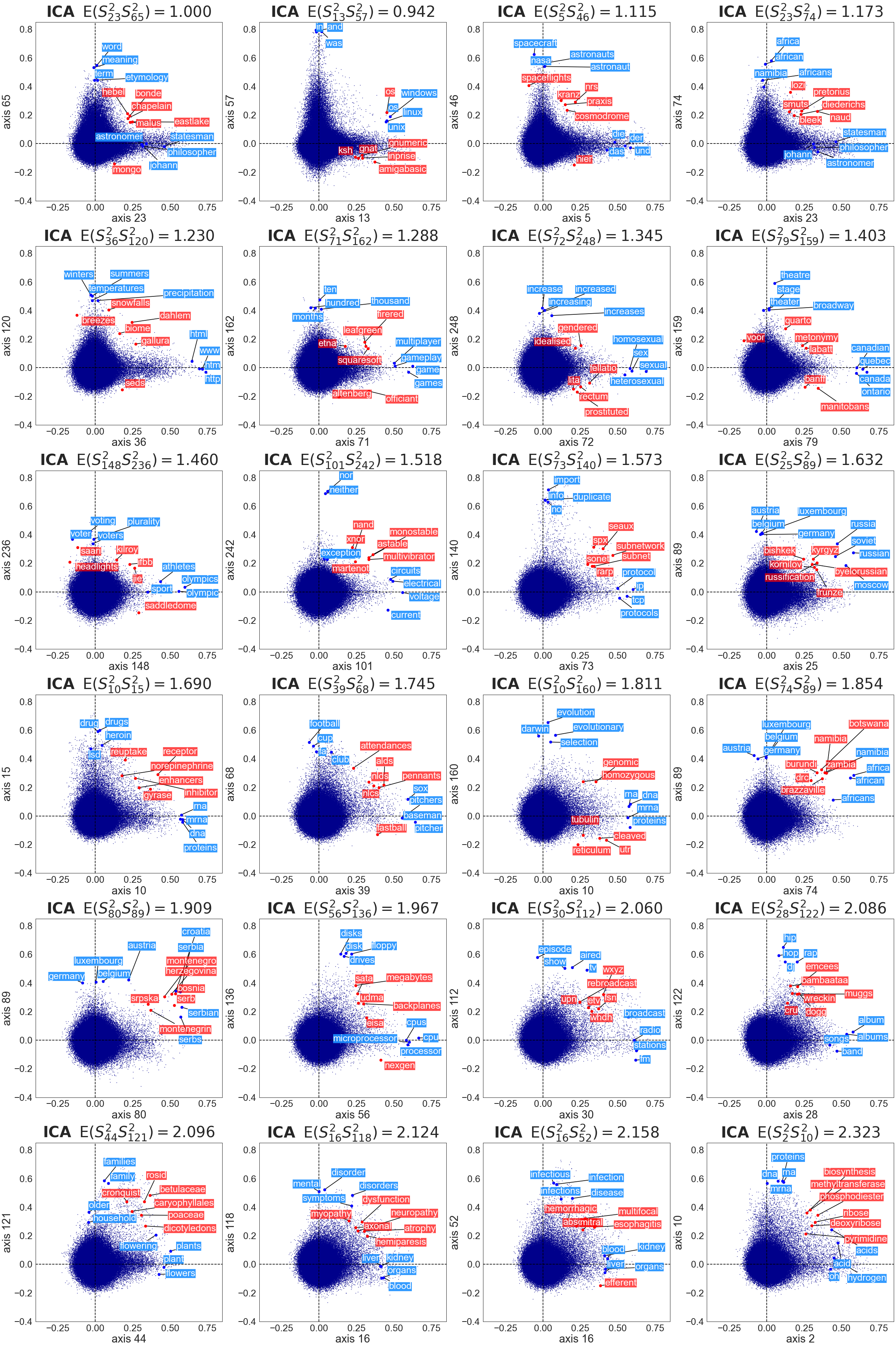}
    \caption{Scatter plots of normalized word embeddings for 24 axis pairs. Blue-labeled words are the top 4 words for each axis's component values, while red-labeled words are the top 6 words for values of $\mathbf{S}_{t,i}^{2} \mathbf{S}_{t,j}^{2}$.}
    \label{fig:scatters24}
\end{figure*}

\paragraph{Complementary Results for Sec.~\ref{subsec:words_example}.}
Table~\ref{tab:xy_words} in Sec.~\ref{subsec:words_example} presented words with significant contributions to higher-order correlations for six axis pairs.
While the main text illustrated the distribution of these highly contributing words through scatter plots for the two selected pairs, Figure~\ref{fig:table3_appendix} provides scatter plots for all the six pairs.

\paragraph{The Relationship Between the Magnitude of Higher-Order Correlations and the Appearance of Scatter Plots.}
Figure~\ref{fig:scatters24} presents the scatter plots of word embeddings for 24 component pairs, each with different higher-order correlation values.
We can see that as the magnitude of higher-order correlation increases, the number of words with large component values along both axes increases as well.
The selection of these 24 pairs was conducted as follows: First, we considered 150 components $S_{\sigma(0)}, \cdots, S_{\sigma(149)}$ with high semantic consistency (see Appendix~\ref{appendix:intrusion} for the calculation method).
We then sorted all possible component pairs $(S_i, S_j)$ $(i, j \in { \sigma(0), \cdots, \sigma(149)})$ based on the value of $|\mathrm{E}(S_i^2 S_j^2) - 1|$.
We established 24 equally spaced grids between the minimum and maximum values, and selected pairs closest to each grid point without repetition.

\paragraph{Words with Significant Contributions to Higher-Order Correlations.}
We have observed words with significant contributions, i.e., with large values of $\mathbf{S}_{t,i}^{2} \mathbf{S}_{t,j}^{2}$, to higher-order correlations $\mathrm{E}(S_{i}^2 S_{j}^2)$ in Table~\ref{tab:xy_words} in Sec.~\ref{subsec:words_example}, and will see further examples in Tables~\ref{tab:subgraph1_alledges} and~\ref{tab:subgraph2_alledges} in Appendix~\ref{appendix:experiment_results}.
Such words are labeled in red in the scatter plots in
Fig.~\ref{fig:table3_maintext} in 
Section~\ref{subsec:words_example}, as well as in
Figs.~\ref{fig:table3_appendix} and \ref{fig:scatters24} in Appendix~\ref{app:scatters_hoc}.
For axis pairs with large higher-order correlations, we observe a large number of words that make significant contributions to the higher-order correlations.
The meanings of these words include both axes' meanings, demonstrating the additive compositionality of embeddings.
 
\section{Prompt Used for Evaluation by GPT-4o mini} \label{appendix:gpt-evaluation}

The specific prompt used for the GPT-4o mini model is provided below.

\begin{lstlisting}
Question:
  You are given 2 list pairs (A, B), (C, D).
  If one pair is more semantically relevant than the other, answer the pair.
  If you cannot determine, answer "XX".

List pair (A, B): ({wordlist_1}, {wordlist_2})
List pair (C, D): ({wordlist_1}, {wordlist_3})

Output:
  "AB" if (A, B) is more semantically related
  "CD" if (C, D) is more semantically related
  "XX" if equally related, or you can't decide
Respond with only AB, CD, or XX.
\end{lstlisting}

When conducting the experiment, we took the following steps to eliminate any potential biases arising from the order of word sequences and specific label names.
\begin{enumerate}
    \item To remove the influence of word order within lists, we randomly shuffled the words in \texttt{wordlist\_1}, \texttt{wordlist\_2}, and \texttt{wordlist\_3}.
    \item To prevent bias in output labels, we conducted the experiment twice, swapping \texttt{wordlist\_2} and \texttt{wordlist\_3} between runs.
    \item To account for order bias in the prompt, we randomly alternated the order of the following two lines:
        \begin{lstlisting}[basicstyle=\tiny\ttfamily]
List pair (A, B): ({wordlist_1}, {wordlist_2})
List pair (C, D): ({wordlist_1}, {wordlist_3})
        \end{lstlisting}
\end{enumerate}

\section{Details of Visualization of Non-Independence Structure} \label{appendix:visualize}

\subsection{Scoring ICA Axes: Word Intrusion Task}\label{appendix:intrusion}
We assigned a semantic coherence score to each axis of the ICA-transformed embeddings using the word intrusion task method~\cite{chang-2009-reading}.

\paragraph{Word Intrusion Task.}
The word intrusion task is a method used to evaluate the semantic coherence of a set of $k$ words by assessing the ability to identify an intruder word. 
For instance, consider the set of words $\{$\emph{windows, os, unix, linux, microsoft}$\}$, which has a consistent theme of operating systems.
In this case, an unrelated word such as \emph{waterskiing} should be easily identifiable as an intruder, as it does not align with the theme of operating systems.
In our experiment, we assigned coherence scores to the top $k=5$ words (with frequency $n_w \geq 100$ in the text8 corpus) for each axis.

\paragraph{Selection of the Intruder Word.} 
In order to select the intruder word for the set of top $k$ words of each axis $a\in\{1,\ldots,d\}$, denoted as $\mathrm{top}_{k}(a)$, we randomly chose a word from a pool of words that satisfy both of the following criteria simultaneously: (i) the word ranks in the lower $50\%$ in terms of the component value on the axis $a$, and (ii) it ranks in the top $10\%$ in terms of the component value on some axis other than $a$.
For each axis, $L=100$ intruder words are randomly selected, and $W_{\mathrm{int}}(a)$ denotes the set of these $L$ intruder words.

\paragraph{Scoring Method.}
For the consistency score of the meaning of each axis $a$, $\mathrm{Score}(a)$, we adopted the metric proposed by \citet{sun-2016-sparse}.
\begin{flalign*}
\mathrm{Score}(a) = \frac{\mathrm{InterDist}(a)}{\mathrm{IntraDist}(a)}  &&
\end{flalign*}
\begin{flalign*}
    \mathrm{IntraDist}(a) &= \sum_{\substack{w_i, w_j \in \mathrm{top}_{k}(a) \\ w_i \neq w_j}} \frac{\mathrm{dist}(w_i, w_j)}{k(k-1)} &&\\
    \mathrm{InterDist}(a) &= \underset{w \in W_{\mathrm{int}}(a)}{\mathrm{mean}} \sum_{w_i \in \mathrm{top}_{k}(a)} \frac{\mathrm{dist}(w_i, w)}{k} &&
\end{flalign*}
In this formula, we defined $\mathrm{dist}(w_i, w_j) = \| \mathbf{s}_i - \mathbf{s}_j\|$ for the ICA-transformed embeddings.
Here, $\mathrm{IntraDist}(a)$ denotes the average distance between the top $k$ words, and $\mathrm{InterDist}(a)$ represents the average distance between the top words and the intruder words. The score is higher when the intruder words are further away from the set $\mathrm{top}_{k}(a)$. Therefore, this score serves as a quantitative measure of the ability to identify the intruder word, thus it is used as a measure of the consistency of the meaning of the top $k$ words and the interpretability of axes.

\subsection{Entire Visualization of MST} \label{appendix:mst}
\begin{figure*}[ht]
    \centering
    \includegraphics[width=0.90\linewidth]{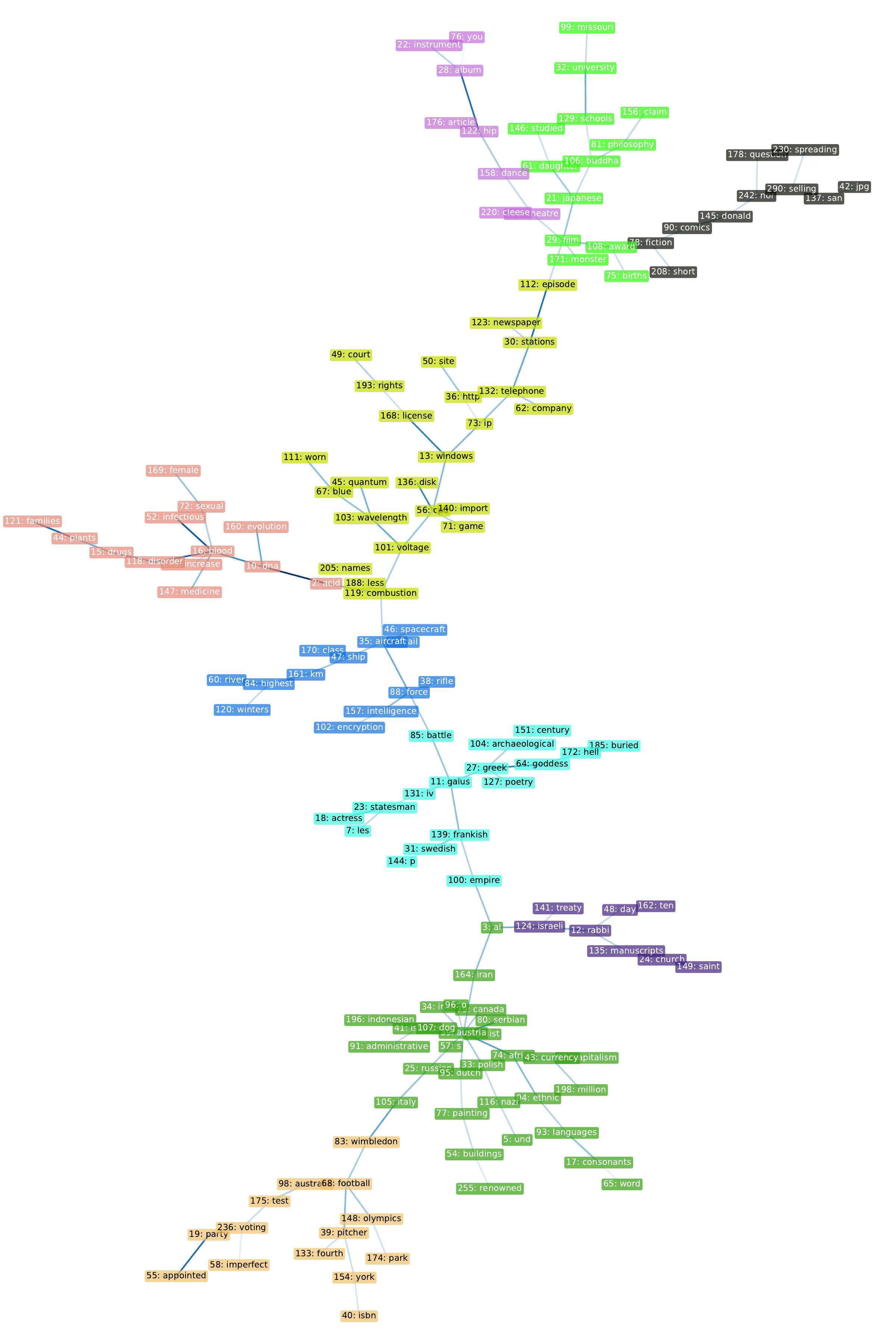}
    \caption{Visualization of the entire MST $T_{150}$ defined in Sec.~\ref{sec:visualization}. Each color of the nodes represents one of the ten clusters obtained by Spectral Clustering.}
    \label{fig:mst_150}
\end{figure*}

Figure~\ref{fig:mst_150} is the visualization of maximum spanning tree (MST) $T_{150}$ defined in Sec.~\ref{sec:visualization}.
For a graph $G_{150}$, where the cost between nodes $i$ and $j$ defined as $c_{ij} = E(S_{i}^{2} S_{j}^{2})$, the algorithm to find the MST $T$ is a greedy method that maximizes the total sum of costs, $\sum_{(i, j) \in T} c_{ij}$, subject to $T$ being a spanning tree.
The greedy algorithm selects edges in decreasing order of $c_{ij}$ while adhering to the tree constraint.
Due to the monotonicity of $f(x) = 1/x$, the decreasing order of $c_{ij}$ is equivalent to the increasing order of $1/c_{ij}$. Thus, computing the MST $T_{150}$ in the graph $G_{150}$ is equivalent to finding the minimum spanning tree, which minimizes the sum of $1/c_{ij}$.

\section{Dimensionality Reduction via MST} \label{appendix:dimension-compression}

\begin{table}[ht]
    \centering
    \begin{adjustbox}{width=\linewidth}
    \begin{tabular}{lcccccc}
    \toprule
         & $d=2$ & $d=5$ & $d=10$ & $d=20$ & $d=50$ & $d=100$\\
        \midrule
        Random Clustering on components (PCA) & 0.04 & 0.08 & 0.12 & 0.16 & 0.23 & 0.29\\
        Random Clustering on components (ICA) & 0.04 & 0.08 & 0.12 & 0.17 & 0.24 & 0.29 \\
        Spectral Clustering on MST (PCA)   & 0.03 & 0.08 & 0.13 & 0.17 & 0.24 & 0.30\\
        Spectral Clustering on MST (ICA)   & \textbf{0.06} & \textbf{0.13} & \textbf{0.18} & \textbf{0.23} & \textbf{0.28} & \textbf{0.31}\\
    \bottomrule
    \end{tabular}
    \end{adjustbox}
    \caption{Word similarity scores for dimensionality reduction.}
    \label{tab:result_dimension_compression}
\end{table}

We experimentally confirmed that the structure of higher-order correlations between components can be applied to the dimensionality reduction of embeddings.
Specifically, we performed Spectral Clustering on the maximum Spanning Tree (MST) $T_{300}$, which was computed based on the higher-order correlations between components. By reducing the dimensionality of the embeddings through averaging the clustered axes, we verified that the accuracy degradation in Word Similarity Tasks was mitigated compared to random clustering.

\paragraph{Experimental Settings.}
We conducted our experiments using the 300-dimensional word embeddings (SGNS). These embeddings were subjected to PCA and ICA to obtain components for clustering. We employed two clustering methods: (1) Random Clustering and (2) Spectral Clustering on the maximum Spanning Tree (MST) $T_{300}$, which was computed based on the higher-order correlations calculated using Eq.~\ref{eq:quadratic-interaction}.
Clustering was performed with the number of clusters ranging from 2 to 100. Dimensionality reduction was achieved by averaging the clustered axes, resulting in reduced dimensions from $d=2$ to $d=100$. The performance of the lower-dimensional embeddings was evaluated through Word Similarity Tasks.
For the Word Similarity Tasks, we utilized six datasets: 
MEN~\cite{ws-MEN}, WS353~\cite{ws-WS353}, MTurk~\cite{ws-MTurk}, RW~\cite{ws-RW}, SimLex999~\cite{ws-SimLex999}, and SimVerb-3500~\cite{ws-SimVerb-3500}.
Each dataset comprises word pairs along with gold similarity scores, assigned by human annotators.
We employed the Spearman rank correlation coefficient between human ratings and the cosine similarity of the word embeddings as the evaluation metric.
The reported values represent the average scores across the six datasets.

\paragraph{Results and Discussion}
The experimental results are presented in Table~\ref{tab:result_dimension_compression}. 
We observe that ICA-based methods  outperform PCA-based methods. 
Moreover, our proposed method, Spectral Clustering on the MST of ICA components, consistently achieves the best performance across all dimensions.
This can be attributed to the fact that components included in the same cluster on the MST likely have high semantic relevance and play similar roles in representing the meaning of words.
These results validate that considering higher-order correlations between axes better preserves semantic relationships in compressed word embeddings, demonstrating the practical utility of our method.
It is important to note that this evaluation assesses the performance of clustering using a downstream task of dimensionality reduction, rather than dimensionality reduction itself.

\section{Supplementary Tables for ICA Components and MST Subtrees} \label{appendix:experiment_results}

Table~\ref{tab:all_topwords} shows all components of the ICA-transformed word embeddings used in our experiments.
Associated with the experiments in Sec.~\ref{subsec:pair_example}, Table~\ref{tab:top60examples} shows the top 60 pairs with the highest \Esisj{} values.
Additionally, in Table~\ref{tab:subgraph1_alledges} and Table~\ref{tab:subgraph2_alledges},
we report on all component pairs in the subtrees of MST $T_{150}$ shown in Fig.~\ref{fig:mst},
extending the results from the selected pairs previously reported in Sec.~\ref{subsec:words_example}.

\begin{table*}[ht]
\centering
\begin{adjustbox}{width=\linewidth}
\begin{tabular}{lllllllllllllllllllllllllllll}
\toprule

Axis $0$ & Axis $1$ & Axis $2$ & Axis $3$ & Axis $4$ & Axis $5$ & Axis $6$ & Axis $7$ & Axis $8$ & Axis $9$ & Axis $10$ & Axis $11$ & Axis $12$ & Axis $13$ & Axis $14$\\
\cmidrule(lr){1-1}\cmidrule(lr){2-2}\cmidrule(lr){3-3}\cmidrule(lr){4-4}\cmidrule(lr){5-5}\cmidrule(lr){6-6}\cmidrule(lr){7-7}\cmidrule(lr){8-8}\cmidrule(lr){9-9}\cmidrule(lr){10-10}\cmidrule(lr){11-11}\cmidrule(lr){12-12}\cmidrule(lr){13-13}\cmidrule(lr){14-14}\cmidrule(lr){15-15}dishes & genus & acid & al & india & und & el & les & chinese & di & dna & gaius & rabbi & windows & topological\\
sauce & species & hydrogen & ibn & indian & der & spanish & du & china & italian & proteins & caesar & talmud & os & isomorphic\\
fried & extinct & acids & muhammad & nehru & die & nacional & des & beijing & della & rna & augustus & rabbis & unix & banach\\
dish & birds & oh & abu & hindu & das & jos & paris & pinyin & luigi & mrna & lucius & torah & linux & topology\\
\cmidrule(lr){1-15}
Axis $15$ & Axis $16$ & Axis $17$ & Axis $18$ & Axis $19$ & Axis $20$ & Axis $21$ & Axis $22$ & Axis $23$ & Axis $24$ & Axis $25$ & Axis $26$ & Axis $27$ & Axis $28$ & Axis $29$\\
\cmidrule(lr){1-1}\cmidrule(lr){2-2}\cmidrule(lr){3-3}\cmidrule(lr){4-4}\cmidrule(lr){5-5}\cmidrule(lr){6-6}\cmidrule(lr){7-7}\cmidrule(lr){8-8}\cmidrule(lr){9-9}\cmidrule(lr){10-10}\cmidrule(lr){11-11}\cmidrule(lr){12-12}\cmidrule(lr){13-13}\cmidrule(lr){14-14}\cmidrule(lr){15-15}drugs & blood & consonants & actress & party & stars & japanese & instrument & statesman & church & russian & cars & greek & album & film\\
drug & organs & vowels & footballer & parties & constellation & japan & instruments & astronomer & churches & moscow & ford & greece & albums & films\\
\bottomrule
 \end{tabular}
 \end{adjustbox}
 \caption{The top $4$ words with the largest component values along all axes of ICA-transformed word embedding used in our experiments. Axes are sorted in descending order of skewness.}
\label{tab:all_topwords}
\end{table*}

\begin{table*}[ht]
\centering
\begin{adjustbox}{width=\linewidth}
 \begin{tabular}{llllllllllllllllllll}
  \toprule

\multicolumn{2}{l}{$\mathrm{E}(S_{63}^2 S_{210}^2) = 2.964$} & \multicolumn{2}{l}{$\mathrm{E}(S_{2}^2 S_{114}^2) = 2.480$} & \multicolumn{2}{l}{$\mathrm{E}(S_{10}^2 S_{125}^2) = 2.431$} & \multicolumn{2}{l}{$\mathrm{E}(S_{58}^2 S_{70}^2) = 2.395$} & \multicolumn{2}{l}{$\mathrm{E}(S_{14}^2 S_{113}^2) = 2.380$} & \multicolumn{2}{l}{$\mathrm{E}(S_{63}^2 S_{184}^2) = 2.330$}\\
Axis 63 & Axis 210 & Axis 2 & Axis 114 & Axis 10 & Axis 125 & Axis 58 & Axis 70 & Axis 14 & Axis 113 & Axis 63 & Axis 184\\
\cmidrule(lr){1-2}\cmidrule(lr){3-4}\cmidrule(lr){5-6}\cmidrule(lr){7-8}\cmidrule(lr){9-10}\cmidrule(lr){11-12}
organization & unesco & acid & morphisms & dna & algae & imperfect & accusative & topological & frac & organization & g\\
international & itu & hydrogen & homomorphism & proteins & bacteria & perfect & nouns & isomorphic & cos & international & e\\
organizations & interpol & acids & hydrogen & rna & fungi & future & genitive & banach & equation & organizations & icao\\
interpol & observer & oh & wavelengths & mrna & mitochondria & present & noun & topology & euler & interpol & fao\\
\midrule[\heavyrulewidth] 
\multicolumn{2}{l}{$\mathrm{E}(S_{2}^2 S_{10}^2) = 2.323$} & \multicolumn{2}{l}{$\mathrm{E}(S_{22}^2 S_{59}^2) = 2.247$} & \multicolumn{2}{l}{$\mathrm{E}(S_{26}^2 S_{134}^2) = 2.233$} & \multicolumn{2}{l}{$\mathrm{E}(S_{14}^2 S_{114}^2) = 2.228$} & \multicolumn{2}{l}{$\mathrm{E}(S_{2}^2 S_{53}^2) = 2.165$} & \multicolumn{2}{l}{$\mathrm{E}(S_{16}^2 S_{52}^2) = 2.158$}\\
Axis 2 & Axis 10 & Axis 22 & Axis 59 & Axis 26 & Axis 134 & Axis 14 & Axis 114 & Axis 2 & Axis 53 & Axis 16 & Axis 52\\
\bottomrule
 \end{tabular}
 \end{adjustbox}
  \caption{Complementary experimental results to Table~\ref{tab:example_energy}. The top 60 pairs with the highest \Esisj{} values are presented. For each component, the top 4 words with the largest component values are listed.}
\label{tab:top60examples}
\end{table*}

\begin{table*}[ht]
\centering
\begin{adjustbox}{width=\linewidth}
 \begin{tabular}{lrlrlrlrlrlrlrlrlrlrlrlr}
  \toprule

\multicolumn{2}{l}{$\mathrm{E}(S_{2}^2 S_{10}^2) = 2.323$} & \multicolumn{2}{l}{$\mathrm{E}(S_{2}^2 S_{119}^2) = 1.755$} & \multicolumn{2}{l}{$\mathrm{E}(S_{132}^2 S_{30}^2) = 1.820$} & \multicolumn{2}{l}{$\mathrm{E}(S_{132}^2 S_{73}^2) = 1.693$} & \multicolumn{2}{l}{$\mathrm{E}(S_{132}^2 S_{62}^2) = 1.632$} & \multicolumn{2}{l}{$\mathrm{E}(S_{136}^2 S_{56}^2) = 1.967$}\\
Axis $2$ & \ml{Axis $10$} & Axis $2$ & \ml{Axis $119$} & Axis $132$ & \ml{Axis $30$} & Axis $132$ & \ml{Axis $73$} & Axis $132$ & \ml{Axis $62$} & Axis $136$ & \ml{Axis $56$}\\
\cmidrule(lr){1-1}\cmidrule(lr){2-2}\cmidrule(lr){3-3}\cmidrule(lr){4-4}\cmidrule(lr){5-5}\cmidrule(lr){6-6}\cmidrule(lr){7-7}\cmidrule(lr){8-8}\cmidrule(lr){9-9}\cmidrule(lr){10-10}\cmidrule(lr){11-11}\cmidrule(lr){12-12}
acid & \ml{dna} & acid & \ml{combustion} & telephone & \ml{stations} & telephone & \ml{ip} & telephone & \ml{company} & disk & \ml{cpu}\\
hydrogen & \ml{proteins} & hydrogen & \ml{diesel} & phone & \ml{fm} & phone & \ml{tcp} & phone & \ml{corporation} & floppy & \ml{microprocessor}\\
acids & \ml{rna} & acids & \ml{turbine} & mobile & \ml{radio} & mobile & \ml{protocols} & mobile & \ml{companies} & disks & \ml{processor}\\
oh & \ml{mrna} & oh & \ml{engine} & cellular & \ml{broadcast} & cellular & \ml{protocol} & cellular & \ml{shareholders} & drives & \ml{cpus}\\
\cmidrule(lr){1-2}\cmidrule(lr){3-4}\cmidrule(lr){5-6}\cmidrule(lr){7-8}\cmidrule(lr){9-10}\cmidrule(lr){11-12}
$w_k$ & $\mathbf{S}_{k,2}^{2}\mathbf{S}_{k,{10}}^{2}$ & $w_k$ & $\mathbf{S}_{k,2}^{2}\mathbf{S}_{k,{119}}^{2}$ & $w_k$ & $\mathbf{S}_{k,132}^{2}\mathbf{S}_{k,{30}}^{2}$ & $w_k$ & $\mathbf{S}_{k,132}^{2}\mathbf{S}_{k,{73}}^{2}$ & $w_k$ & $\mathbf{S}_{k,132}^{2}\mathbf{S}_{k,{62}}^{2}$ & $w_k$ & $\mathbf{S}_{k,136}^{2}\mathbf{S}_{k,{56}}^{2}$\\
\bottomrule
 \end{tabular}
 \end{adjustbox}
 \caption{Complementary experimental results to Table~\ref{tab:xy_words}. For all component pairs \sisj{} in the first subtree of the MST in Fig.~\ref{fig:mst}, the top 6 words and their corresponding $\mathbf{S}_{t,i}^{2} \mathbf{S}_{t,j}^{2}$ values that contribute the most to the \Esisj{} value are presented.}
\label{tab:subgraph1_alledges}
\end{table*}

\begin{table*}[ht]
\centering
\begin{adjustbox}{width=\linewidth}
 \begin{tabular}{lrlrlrlrlrlrlrlrlrlrlrlr}
  \toprule

\multicolumn{2}{l}{$\mathrm{E}(S_{2}^2 S_{10}^2) = 2.323$} & \multicolumn{2}{l}{$\mathrm{E}(S_{2}^2 S_{119}^2) = 1.755$} & \multicolumn{2}{l}{$\mathrm{E}(S_{132}^2 S_{30}^2) = 1.820$} & \multicolumn{2}{l}{$\mathrm{E}(S_{132}^2 S_{73}^2) = 1.693$} & \multicolumn{2}{l}{$\mathrm{E}(S_{132}^2 S_{62}^2) = 1.632$} & \multicolumn{2}{l}{$\mathrm{E}(S_{136}^2 S_{56}^2) = 1.967$}\\
Axis $2$ & \ml{Axis $10$} & Axis $2$ & \ml{Axis $119$} & Axis $132$ & \ml{Axis $30$} & Axis $132$ & \ml{Axis $73$} & Axis $132$ & \ml{Axis $62$} & Axis $136$ & \ml{Axis $56$}\\
\cmidrule(lr){1-1}\cmidrule(lr){2-2}\cmidrule(lr){3-3}\cmidrule(lr){4-4}\cmidrule(lr){5-5}\cmidrule(lr){6-6}\cmidrule(lr){7-7}\cmidrule(lr){8-8}\cmidrule(lr){9-9}\cmidrule(lr){10-10}\cmidrule(lr){11-11}\cmidrule(lr){12-12}
acid & \ml{dna} & acid & \ml{combustion} & telephone & \ml{stations} & telephone & \ml{ip} & telephone & \ml{company} & disk & \ml{cpu}\\
hydrogen & \ml{proteins} & hydrogen & \ml{diesel} & phone & \ml{fm} & phone & \ml{tcp} & phone & \ml{corporation} & floppy & \ml{microprocessor}\\
acids & \ml{rna} & acids & \ml{turbine} & mobile & \ml{radio} & mobile & \ml{protocols} & mobile & \ml{companies} & disks & \ml{processor}\\
oh & \ml{mrna} & oh & \ml{engine} & cellular & \ml{broadcast} & cellular & \ml{protocol} & cellular & \ml{shareholders} & drives & \ml{cpus}\\
\cmidrule(lr){1-2}\cmidrule(lr){3-4}\cmidrule(lr){5-6}\cmidrule(lr){7-8}\cmidrule(lr){9-10}\cmidrule(lr){11-12}
$w_k$ & $\mathbf{S}_{k,2}^{2}\mathbf{S}_{k,{10}}^{2}$ & $w_k$ & $\mathbf{S}_{k,2}^{2}\mathbf{S}_{k,{119}}^{2}$ & $w_k$ & $\mathbf{S}_{k,132}^{2}\mathbf{S}_{k,{30}}^{2}$ & $w_k$ & $\mathbf{S}_{k,132}^{2}\mathbf{S}_{k,{73}}^{2}$ & $w_k$ & $\mathbf{S}_{k,132}^{2}\mathbf{S}_{k,{62}}^{2}$ & $w_k$ & $\mathbf{S}_{k,136}^{2}\mathbf{S}_{k,{56}}^{2}$\\
\bottomrule
 \end{tabular}
 \end{adjustbox}
 \caption{Complementary experimental results to Table~\ref{tab:xy_words}. For all component pairs \sisj{} in the second subtree of the MST in Fig.~\ref{fig:mst}, the top 6 words and their corresponding $\mathbf{S}_{t,i}^{2} \mathbf{S}_{t,j}^{2}$ values that contribute the most to the \Esisj{} value are presented.}
\label{tab:subgraph2_alledges}
\end{table*}
\end{document}